\documentclass[11pt]{article}

\usepackage{microtype}
\usepackage{graphicx}
\usepackage{subcaption}
\usepackage{booktabs}
\usepackage{microsoft-tech-report}
\usepackage{hyperref}
\usepackage{url}
\usepackage[utf8]{inputenc}
\usepackage[T1]{fontenc}
\usepackage{amsmath}
\usepackage{amssymb}
\usepackage{amsfonts}
\usepackage{mathtools}
\usepackage{amsthm}
\usepackage{bm}
\usepackage{nicefrac}
\usepackage{enumitem}
\usepackage{multirow}
\usepackage[capitalize,noabbrev]{cleveref}
\usepackage{xspace}
\usepackage{pifont}
\usepackage{wrapfig}
\usepackage{algorithm}
\usepackage{algpseudocode}
\usepackage{placeins}

\newcommand{\ourmethod}{\textsc{SkillOpt}\xspace}
\newcommand{\na}{--}

\definecolor{deltapos}{HTML}{0E8A4A}
\definecolor{deltaneg}{HTML}{B91C1C}

\newcommand{\dvalp}[2]{\hphantom{\textsubscript{+#2}}#1\textsubscript{\textcolor{deltapos}{+#2}}}
\newcommand{\dvaln}[2]{\hphantom{\textsubscript{$-$#2}}#1\textsubscript{\textcolor{deltaneg}{$-$#2}}}

\techreportlabel{Microsoft Research}
\techreportshorttitle{SkillOpt}

\hypersetup{
  colorlinks=true,
  linkcolor=msftblue,
  citecolor=msftblue,
  urlcolor=msftblue,
  pdftitle={SkillOpt: Executive Strategy for Self-Evolving Agent Skills}
}

\begin{document}
\thispagestyle{empty}

\noindent
\begin{minipage}[c]{0.5\linewidth}
\raggedright
\raisebox{-0.5\height}{\msftbrandmark}
\end{minipage}%
\begin{minipage}[c]{0.49\linewidth}
\raggedleft
{\msftdatefont\small\color{msftgray}May 2026}
\end{minipage}\par
\vspace{0.35em}
\noindent{\color{msftline}\rule{\linewidth}{0.8pt}\par}

\vspace{1.0em}
\begin{center}
{{\msfttitlefont\fontsize{21}{25}\selectfont\color{msftdark}
\ourmethod{}: Executive Strategy for\\
Self-Evolving Agent Skills\par}}
\vspace{1.25em}

{\normalsize\rmfamily\color{msftdark}
Yifan Yang$^{1,*,\ddagger}$ \hspace{0.6em}
Ziyang Gong$^{2,*}$ \hspace{0.6em}
Weiquan Huang$^{3,*}$ \hspace{0.6em}
Qihao Yang$^{2,*}$ \hspace{0.6em}
Ziwei Zhou$^{4,*}$\\[-0.1em]
Zisu~Huang$^{4,*}$ \hspace{0.6em}
Yan Li$^{2}$ \hspace{0.6em}
Xuemei Gao$^{1}$ \hspace{0.6em}
Qi Dai$^{1}$ \hspace{0.6em}
Bei Liu$^{1}$\\[-0.1em]
Kai Qiu$^{1}$ \hspace{0.6em}
Yuqing Yang$^{1}$ \hspace{0.6em}
Dongdong Chen$^{1}$ \hspace{0.6em}
Xue Yang$^{2,\ddagger}$ \hspace{0.6em}
Chong Luo$^{1}$\par
}
\vspace{0.22cm}

{\footnotesize\rmfamily\color{msftgray}
$^{1}$ Microsoft \quad
$^{2}$ Shanghai Jiao Tong University \quad
$^{3}$ Tongji University \quad
$^{4}$ Fudan University\par
}
\end{center}

\vspace{0.45em}
\begin{msfttitlebox}
\setlength{\parindent}{0cm}
\setlength{\parskip}{0.14cm}
\raggedright
\nohyphens

Agent skills today are hand-crafted, generated one-shot, or evolved through loosely controlled self-revision---none of which behaves like a deep-learning optimizer for the skill, and none of which reliably improves over its starting point under feedback. We argue the skill should instead be \emph{trained} as the external state of a frozen agent, with the same discipline that makes weight-space optimization reproducible. \ourmethod{} is, to our knowledge, the first systematic \emph{controllable} text-space optimizer for agent skills: a separate optimizer model turns scored rollouts into bounded add/delete/replace edits on a single skill document, and an edit is accepted only when it strictly improves a held-out validation score. A textual learning-rate budget, rejected-edit buffer, and epoch-wise slow/meta update make skill training stable while adding zero inference-time model calls at deployment. Across six benchmarks, seven target models, and three execution harnesses (direct chat, Codex, Claude Code), \ourmethod{} is best or tied on \textbf{all 52 evaluated (model, benchmark, harness) cells} and beats every per-cell competitor among human, one-shot LLM, Trace2Skill, TextGrad, GEPA, and EvoSkill skills. On GPT--5.5 it lifts the average no-skill accuracy by $\mathbf{+23.5}$ points in direct chat, by $+24.8$ inside the Codex agentic loop, and by $+19.1$ inside Claude Code. Transfer experiments further show that optimized skill artifacts retain value when moved across model scales, between Codex and Claude Code execution environments, and to a nearby math benchmark without further optimization.

\vspace{0.14cm}
{\setlength{\parskip}{0.06cm}\small
{\msftmetalabel{Code}\href{https://aka.ms/SkillOpt}{https://aka.ms/SkillOpt}\par}
{\msftmetalabel{Correspondence}
\href{mailto:yifanyang@microsoft.com}{yifanyang@microsoft.com},
\href{mailto:yangxue2019-sjtu@sjtu.edu.cn}{yangxue2019-sjtu@sjtu.edu.cn}\par}
}
\vspace{0.08cm}
{\footnotesize\rmfamily\itshape\color{msftgray}
$^{*}$ Equal contribution. \quad $^{\ddagger}$ Corresponding authors.\par
}
\end{msfttitlebox}


\section{Introduction}

\begin{figure}[t]
    \centering
    \includegraphics[width=\textwidth]{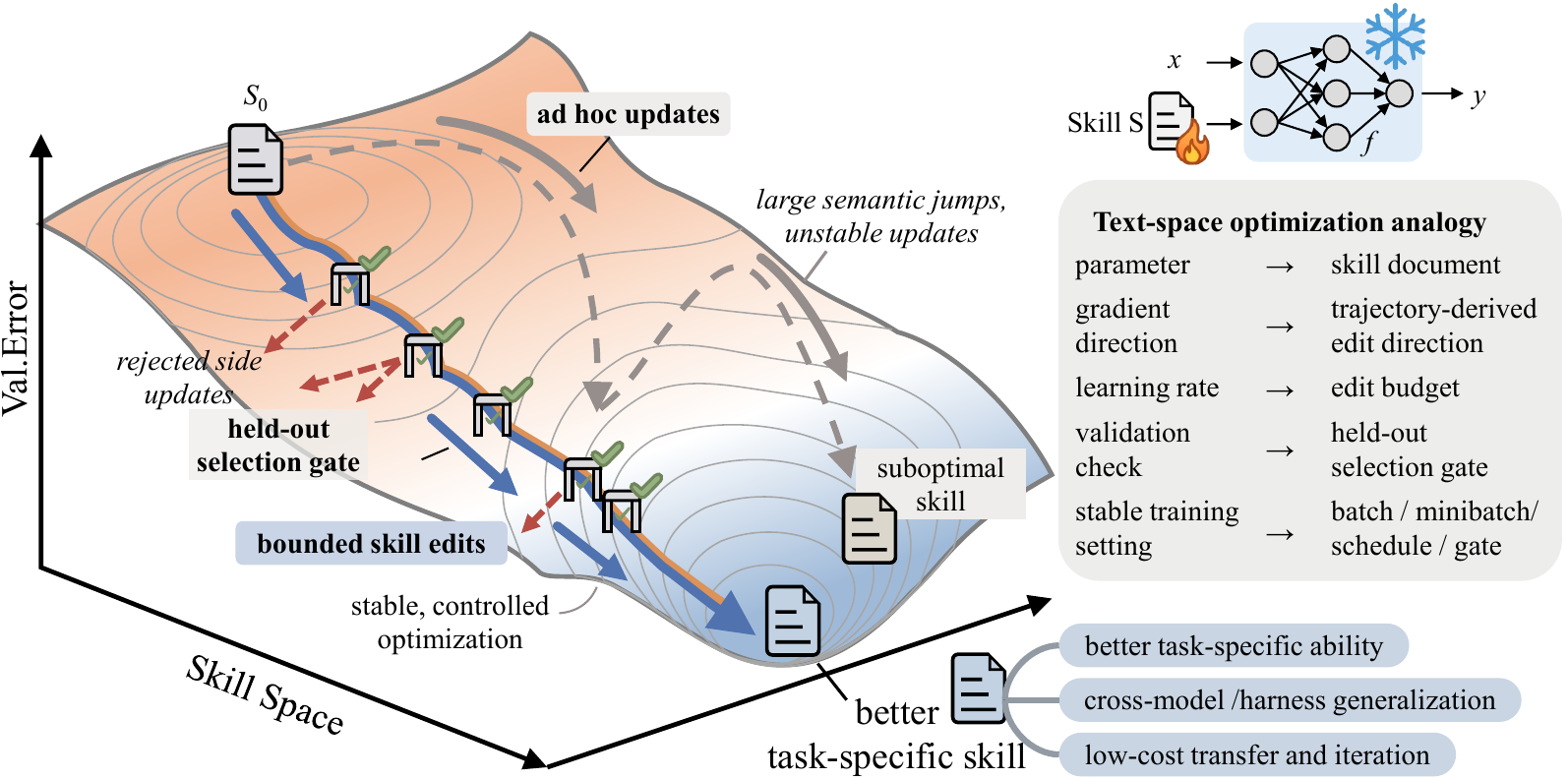}
    \caption{Overview of \ourmethod{}. The target model executes tasks with a current skill, an additional frontier optimizer model converts trajectories into bounded add/delete/replace skill edits, and a held-out gate accepts only edits that improve validation performance. Accepted edits are exported as a reusable skill artifact, while rejected edits become negative feedback for later updates.}
    \label{fig:skillopt_teaser}
\end{figure}

Frontier language models are increasingly deployed as agents, from single-prompt callers to multi-step execution harnesses with tools, files, and verifiers \citep{yao2022react,schick2023toolformer,wang2023voyager,yang2024sweagent}. In such settings, domain adaptation is no longer only about model weights or prompts: it also requires improving the \emph{procedures} by which the agent gathers evidence, calls tools, follows domain conventions, and formats outputs \citep{yang2023large,khattab2023dspy}. Agent skills provide a natural interface for this procedural adaptation \citep{li2026skillsbench,jiang2026sok}: a skill is a portable natural-language artifact that packages procedures, domain heuristics, tool policies, output constraints, and failure modes, letting a frozen agent adapt through external text.

If the recurring object of adaptation is the agent's procedure, the skill document itself should be trainable. Yet weight adaptation is often unavailable for closed frontier models and expensive for open ones, while manually written or one-shot skills are brittle under a target domain or harness. Recent systems convert execution experience into reusable textual artifacts---distilling trajectory lessons, refining skill folders via failure analysis, building domain-specific skill libraries, or optimizing prompts from trajectory feedback \citep{ni2026trace2skill,alzubi2026evoskill,luo2026skillforge,shen2026skillfoundry,agrawal2025gepa}---but leave open a more basic question: if skills are the adaptation layer, how should they be optimized? Our key idea is to treat skill editing as a controllable domain-adaptation process, with the skill document as the external state, an additional frontier model as the optimizer, and training-style controls over evidence, step size, validation, and update direction.

We introduce \ourmethod{}, a text-space optimizer for agent skills. Given a target domain, an initial skill, and the model being adapted, \ourmethod{} repeatedly samples trajectory batches, analyzes successes and failures, and asks a frontier optimizer model to propose structured add/delete/replace edits. It then aggregates and ranks candidate edits under a textual learning-rate budget, applies a bounded update to the skill document, and evaluates the candidate skill on a held-out selection split before accepting it. Rejected edits are retained as negative feedback, while the epoch-wise slow/meta update preserves longer-horizon regularities. Figure~\ref{fig:skillopt_teaser} gives a schematic view of this loop. The deployed output is a compact \mbox{\texttt{best\_skill.md}} file of roughly $300$--$2{,}000$ tokens, with the adapted model and execution harness remaining fixed.

The deep-learning analogy is operational rather than decorative. Rollout and reflection batch sizes control the noise in the evidence used for each edit; the textual learning rate and schedule control how far one skill version is allowed to move from the previous one; the held-out gate plays the role of validation; and the epoch-wise slow/meta update acts like a momentum term, carrying stable editing directions across epochs. This stability is crucial: if consecutive skill revisions move too far or in inconsistent directions, rejected edits and previous accepted edits no longer provide a meaningful optimization history. With bounded, validation-gated updates, each revision remains close enough to the last one that later optimizer calls can learn from what helped, what failed, and what should be preserved.

We conduct, to our knowledge, the first systematic study of skill optimization as a domain-adaptation training method for frontier agents. We evaluate \ourmethod{} on six benchmarks covering QA, spreadsheets, documents, math, and embodied decision making, across seven target models from frontier-scale GPT to small-scale Qwen, and under three execution modes (direct chat, Codex harness, Claude Code harness). Out of 52 evaluated (model, benchmark, harness) cells, \ourmethod{} is the best or tied-best measured method on all 52. With GPT--5.5 in direct chat, it lifts SearchQA from 77.7 to 87.3, SpreadsheetBench from 41.8 to 80.7, OfficeQA from 33.1 to 72.1, DocVQA from 78.8 to 91.2, LiveMathematicianBench from 37.6 to 66.9, and ALFWorld from 83.6 to 95.5 (a $+23.5$ point average gain over no skill), and it also beats the strongest \emph{per-cell} baseline drawn from human-written, one-shot LLM, Trace2Skill, TextGrad, GEPA, and EvoSkill skills by $+5.4$ points on average. The same optimization interface is effective inside Codex-style and Claude Code-style execution loops, lifting GPT--5.5 by $+24.8$ and $+19.1$ points over no skill respectively, and outperforming EvoSkill by $+14.0$ and $+3.2$ points.

The learned artifacts also transfer beyond the exact training setting. A SpreadsheetBench skill trained on GPT--5.4 improves every smaller GPT variant we test; a Codex-trained spreadsheet skill transfers to Claude Code with a $+59.7$ point gain; and an OlympiadBench skill yields positive gains on Omni-MATH~\citep{gao2024omnimathuniversalolympiadlevel}. These transfer results are important for the paper's application value: a skill can be optimized once, audited as text, and reused across related models, harnesses, or tasks without changing model weights. Our ablations explain why this works. Bounded textual learning outperforms uncontrolled rewriting, held-out gating prevents harmful proposals from accumulating, the rejected-step buffer converts failed edits into negative feedback, and the epoch-wise slow/meta update improves long-horizon refinement without bloating the deployed skill. Finally, per-benchmark case studies show that the learned skills remain compact ($300$--$2{,}000$ tokens after only $1$--$4$ accepted edits), inspectable, and procedural rather than instance-specific.

Our contributions are as follows:
\begin{itemize}
    \item We formulate agent-skill learning as optimization over an external natural-language state and introduce \ourmethod{}, a harness-agnostic optimizer with rollout batches, reflection minibatches, add/delete/replace edits, textual learning rates, schedules, held-out acceptance, rejected-edit buffers, and epoch-wise slow/meta update.
    \item We provide a broad empirical study across six benchmarks, seven target models, and three execution harnesses, showing that \ourmethod{} is best or tied-best on 52 of 52 cells and outperforms no-skill, human-skill, one-shot LLM-skill, prompt-optimization (TextGrad, GEPA), and skill-evolution (Trace2Skill, EvoSkill) baselines under every model.
    \item We validate the optimization design through component ablations and three forms of transfer (cross-model, cross-harness, cross-benchmark), showing that the exported skill artifact is compact, reusable, and deployable without model-weight updates.
\end{itemize}

\section{Related Work}
\label{sec:related_work}

\paragraph{Prompt auto tuning and agent-configuration search.}
GEPA demonstrates that trajectory feedback can guide reflective prompt evolution and outperform reinforcement learning on several language-agent tasks \citep{agrawal2025gepa}. ABSTRAL and EvoTest extend this idea from single prompts to multi-agent design documents and test-time agentic system evolution without gradients or fine-tuning \citep{song2026abstral,he2025evotest}. By treating language artifacts as optimizable objects, these methods can directly exploit execution feedback, but they mainly target prompts, system designs, or full configurations rather than reusable domain adaptation. \ourmethod{} instead optimizes a persistent skill document that can be trained, validated, exported, and reused with the adapted model, applying language-level controllability to a stable procedural skill state.

\paragraph{Skill construction and skill evolution.}
SkillsBench and the SoK on agentic skills frame skills as reusable procedural knowledge, covering tool policies, applicability conditions, execution routines, and supporting resources \citep{li2026skillsbench,jiang2026sok}. Prior systems construct such skills from lifelong experience, trajectory lessons, skill knowledge bases, or heterogeneous domain resources \citep{yang2026autoskill,ni2026trace2skill,wang2026skillx,shen2026skillfoundry,memp}, and further refine them through failure analysis, creation-evaluation-revision loops, co-evolving generators and verifiers, collective updates, or reinforcement learning \citep{alzubi2026evoskill,luo2026skillforge,zhang2026coevoskills,ma2026skillclaw,xia2026skillrl,wang2025reinforcement,autorefine,procmem,evolver}. While these works emphasize skill discovery, repository growth, sharing, evolutionary search, or policy optimization, \ourmethod{} studies a narrower problem: how to train one compact domain skill with deep-learning-style controls such as trajectory batches, reflection minibatches, textual learning rates, validation gates, rejected-edit buffers, and slow/meta updates. This yields a controlled and auditable procedure for producing a portable \texttt{best\_skill.md} without changing model weights.

\providecommand{\ourmethod}{\textsc{SkillOpt}}
\providecommand{\na}{--}
\providecommand{\harnessrow}[1]{\rowcolor{black!7}\multicolumn{8}{l}{\textbf{#1}}}
\newcommand{\methodcell}[1]{\cellcolor{blue!8}#1}
\newcommand{\ablationbenchmarkheader}{\textbf{Setting} & \textbf{SearchQA} & \textbf{Spreadsheet} & \textbf{LiveMath} \\}

\section{Method}
\label{sec:method}

\begin{figure}[t]
    \centering
    \includegraphics[width=\textwidth]{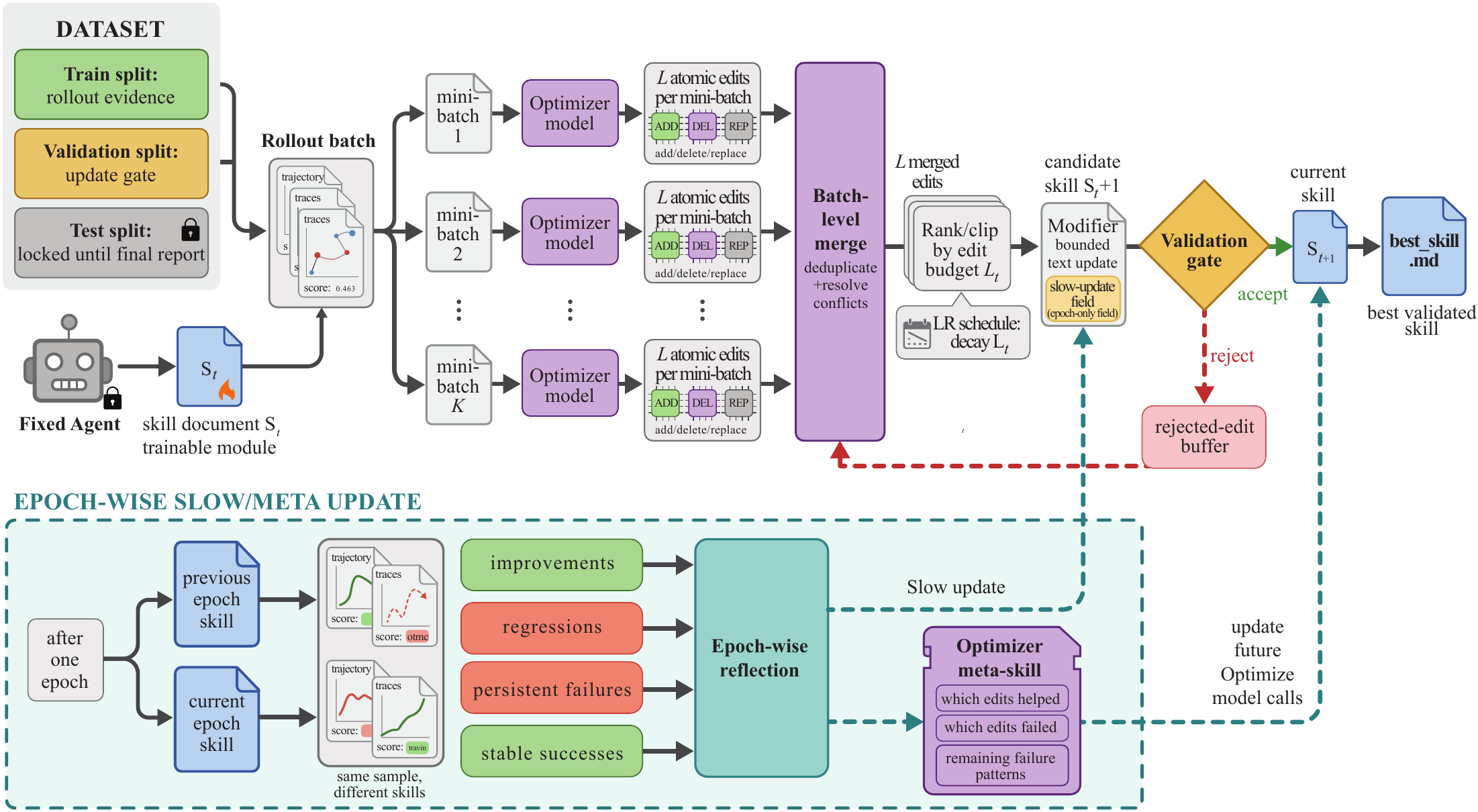}
    \caption{Pipeline of \ourmethod{}. A frozen target model executes a rollout batch with the current skill; an optimizer model performs minibatch reflection over successes and failures, proposes bounded add/delete/replace edits, merges and ranks them under a scheduled edit budget, and accepts the candidate skill only through a held-out validation gate. Across epochs, the slow/meta update retains longer-horizon lessons without changing the target model.}
    \label{fig:skillopt_pipeline}
\end{figure}


\subsection{Problem Setup}
\label{sec:method_skill}

A skill $s$ is a natural-language policy inserted into the agent context before execution, consistent with recent work treating skills as reusable procedural knowledge for agents \citep{li2026skillsbench,jiang2026sok}. In direct-chat benchmarks, it is prepended to the system or developer instruction; in tool-use harnesses, it becomes persistent procedural memory. We use $M$ to denote the frozen target model whose behavior is being adapted through skill optimization. For a harness $h$, task $x$, and skill $s$, execution produces a trajectory $\tau$ and a scalar score $r$:
\begin{equation}
    (\tau(s), r(s)) = h(M, x, s), \qquad r(s) \in [0,1].
\end{equation}

Given train, selection, and test splits $D_{\mathrm{tr}},D_{\mathrm{sel}},D_{\mathrm{test}}$, \ourmethod{} uses $D_{\mathrm{tr}}$ to generate a set of candidate skills $\mathcal{C}(D_{\mathrm{tr}})$, selects the best skill on $D_{\mathrm{sel}}$, and reports the final performance on $D_{\mathrm{test}}$:
\begin{equation}
    s^\star_{\mathrm{sel}}
    =
    \arg\max_{s \in \mathcal{C}(D_{\mathrm{tr}})}
    \frac{1}{|D_{\mathrm{sel}}|}
    \sum_{x \in D_{\mathrm{sel}}}
    r(s),
\end{equation}
\begin{equation}
    \mathrm{Test}(s^\star_{\mathrm{sel}})
    =
    \frac{1}{|D_{\mathrm{test}}|}
    \sum_{x \in D_{\mathrm{test}}}
    r(s^\star_{\mathrm{sel}}).
\end{equation}
The training split supplies experience, the selection split gates updates, and the test split is used only for final reporting. The optimizer state contains the current skill, the best validation-gated skill, cached skill hashes, an epoch-local rejected-step buffer, and optional slow/meta-update state. Only the best accepted skill is exported as \texttt{best\_skill.md}.

\subsection{Forward Pass: Rollout Evidence}
\label{sec:method_forward}

At each optimization step, the target model runs a rollout batch from $D_{\mathrm{tr}}$ with the current skill. The harness records task metadata, messages, tool calls, observations, command outputs, final answers, verifier feedback, and benchmark-specific context such as spreadsheet previews, document references, or compact execution traces. This batch is the evidence unit: small batches update quickly but noisily, while larger batches expose more recurring patterns before the skill changes. The implementation also supports accumulation, where several rollout batches are reflected on separately and merged into one update, decoupling execution throughput from update frequency.

\subsection{Backward Pass: Minibatch Reflection}
\label{sec:method_backward}

The optimizer model turns trajectories into skill edits, following the broader line of trajectory-driven reflection and prompt evolution \citep{shinn2023reflexion,madaan2023selfrefine,agrawal2025gepa}. It first separates failures from successes and partitions each group into reflection minibatches. This matters because single trajectories often produce anecdotal fixes, while minibatches expose reusable procedural errors: the agent consistently searches the wrong source, writes an answer in the wrong format, or fails to verify a tool result. Failure minibatches propose missing or corrective rules; success minibatches preserve behaviors that already work. Each reflection returns structured add/delete/replace edits, or in rewrite mode a small set of rewrite suggestions.

Local proposals are merged hierarchically by first consolidating failure- and success-driven edits separately, then combining them with priority on failure corrections. This step filters duplicate, contradictory, and example-specific suggestions before the optimizer selects the final bounded update.

\subsection{Bounded Text Updates}
\label{sec:method_controls}

The learning-rate analogue in \ourmethod{} is the edit budget $L_t$: the maximum number of skill edits applied at step $t$. After aggregation, the optimizer model ranks the merged edit pool by expected utility and clips it to the top $L_t$ edits. This is the key difference from ad hoc prompt rewriting. Unbounded rewrites can erase useful rules, introduce incompatible instructions, or overfit to a local failure; bounded updates preserve continuity while still allowing the skill to acquire new procedures. \ourmethod{} supports constant, linear, cosine, and autonomous schedules. The default cosine schedule starts with larger edits and decays toward smaller consolidation steps.

The selected edits produce a candidate skill. In patch mode, edits are localized operations such as append, insert, replace, and delete; in rewrite mode, selected suggestions condition a full skill rewrite. Step-level edits cannot overwrite the protected slow-update field, so fast local changes and slower epoch-wise consolidation remain separated.

\subsection{Validation Gate and Rejected-Edit Buffer}
\label{sec:method_gate}

Every candidate skill is evaluated on $D_{\mathrm{sel}}$ with the same frozen target model and harness. If it improves over the current selection score, it becomes the new current skill; if it also exceeds the best score so far, it becomes \texttt{best\_skill.md}. Otherwise it is rejected. This gate turns reflection into propose-and-test optimization rather than unconditional self-editing, which is crucial because plausible textual diagnoses can still hurt the actual target model.

Rejected updates are still useful. The optimizer records an epoch-local buffer containing observed failure patterns and, for rejected steps, the edits that were tried and the score drop they caused. Later reflection calls in the same epoch receive this buffer, so the optimizer model can avoid repeating failed edits and focus on unresolved failures. This gives the loop negative feedback during training without adding inference-time cost.
\begin{table*}[!htbp]
\centering
\scriptsize
\setlength{\tabcolsep}{1.5pt}
\renewcommand{\arraystretch}{1.0}
\begin{tabular}{l l c c c c c c}
\toprule
\textbf{Model} & \textbf{Skill source} & \textbf{SearchQA} & \textbf{Spreadsheet} & \textbf{OfficeQA} & \textbf{DocVQA} & \textbf{LiveMath} & \textbf{ALFWorld} \\
\midrule
\harnessrow{No harness / direct chat} \\
\multirow{7}{*}{GPT--5.5} & No skill & 77.7 & 41.8 & 33.1 & 78.8 & 37.6 & 83.6 \\
 & Human skill & \dvalp{81.8}{4.1} & \dvalp{72.9}{31.1} & \dvalp{\underline{66.9}}{33.8} & \dvalp{90.1}{11.3} & \dvalp{38.4}{0.8} & \dvalp{91.8}{8.2} \\
 & LLM skill & \dvalp{80.9}{3.2} & \dvalp{43.2}{1.4} & \dvalp{51.7}{18.6} & \dvalp{89.6}{10.8} & \dvalp{40.0}{2.4} & \dvalp{\underline{93.3}}{9.7} \\
 & Trace2Skill & \dvalp{82.4}{4.7} & \dvalp{49.6}{7.8} & \dvalp{65.7}{32.6} & \dvalp{\underline{90.6}}{11.8} & \dvalp{\underline{52.0}}{14.4} & \dvalp{87.3}{3.7} \\
 & TextGrad & \dvalp{81.4}{3.7} & \dvaln{41.1}{0.7} & \dvalp{42.0}{8.9} & \dvalp{87.2}{8.4} & \dvalp{49.2}{11.6} & \dvaln{82.8}{0.8} \\
 & GEPA & \dvalp{\underline{84.8}}{7.1} & \dvalp{\underline{73.6}}{31.8} & \dvalp{63.9}{30.8} & \dvalp{89.1}{10.3} & \dvalp{43.2}{5.6} & \dvalp{85.8}{2.2} \\
 & \methodcell{\textbf{\ourmethod{}}} & \methodcell{\dvalp{\textbf{87.3}}{9.6}} & \methodcell{\dvalp{\textbf{80.7}}{38.9}} & \methodcell{\dvalp{\textbf{72.1}}{39.0}} & \methodcell{\dvalp{\textbf{91.2}}{12.4}} & \methodcell{\dvalp{\textbf{66.9}}{29.3}} & \methodcell{\dvalp{\textbf{95.5}}{11.9}} \\
\cmidrule(lr){1-8}
\multirow{7}{*}{GPT--5.4} & No skill & 76.9 & 41.4 & 50.0 & 77.6 & 36.8 & 75.4 \\
 & Human skill & \dvalp{80.1}{3.2} & \dvalp{59.3}{17.9} & \dvalp{59.3}{9.3} & \dvalp{88.5}{10.9} & \dvalp{\underline{41.6}}{4.8} & \dvaln{74.6}{0.8} \\
 & LLM skill & \dvalp{79.4}{2.5} & \dvalp{46.1}{4.7} & \dvaln{20.4}{29.6} & \dvalp{\underline{90.4}}{12.8} & \dvalp{36.8}{0.0} & \dvalp{\underline{83.6}}{8.2} \\
 & Trace2Skill & \dvalp{81.2}{4.3} & \dvalp{53.2}{11.8} & \dvalp{51.2}{1.2} & \dvalp{89.3}{11.7} & \dvalp{40.0}{3.2} & \dvaln{73.1}{2.3} \\
 & TextGrad & \dvalp{\underline{82.5}}{5.6} & \dvaln{38.6}{2.8} & \dvalp{54.7}{4.7} & \dvalp{85.3}{7.7} & \dvaln{36.6}{0.2} & \dvalp{78.4}{3.0} \\
 & GEPA & \dvalp{82.4}{5.5} & \dvalp{\underline{61.1}}{19.7} & \dvalp{\underline{60.4}}{10.4} & \dvalp{88.3}{10.7} & \dvalp{\underline{41.6}}{4.8} & \dvaln{74.6}{0.8} \\
 & \methodcell{\textbf{\ourmethod{}}} & \methodcell{\dvalp{\textbf{83.1}}{6.2}} & \methodcell{\dvalp{\textbf{62.5}}{21.1}} & \methodcell{\dvalp{\textbf{62.8}}{12.8}} & \methodcell{\dvalp{\textbf{91.2}}{13.6}} & \methodcell{\dvalp{\textbf{44.0}}{7.2}} & \methodcell{\dvalp{\textbf{91.0}}{15.6}} \\
\cmidrule(lr){1-8}
\multirow{7}{*}{GPT--5.4-mini} & No skill & 75.9 & 36.1 & 22.1 & 71.4 & 14.7 & 73.1 \\
 & Human skill & \dvalp{77.2}{1.3} & \dvalp{\underline{42.9}}{6.8} & \dvalp{\underline{45.9}}{23.8} & \dvalp{85.0}{13.6} & \dvalp{\underline{28.8}}{14.1} & \dvaln{56.7}{16.4} \\
 & LLM skill & \dvalp{76.1}{0.2} & \dvalp{36.8}{0.7} & \dvalp{36.6}{14.5} & \dvalp{86.4}{15.0} & \dvalp{28.0}{13.3} & \dvaln{65.7}{7.4} \\
 & Trace2Skill & \dvalp{78.6}{2.7} & \dvalp{40.7}{4.6} & \dvaln{20.9}{1.2} & \dvalp{\underline{88.5}}{17.1} & \dvalp{\textbf{32.8}}{18.1} & \dvalp{\underline{82.8}}{9.7} \\
 & TextGrad & \dvalp{77.5}{1.6} & \dvalp{38.2}{2.1} & \dvalp{30.0}{7.9} & \dvalp{84.0}{12.6} & \dvalp{27.2}{12.5} & \dvaln{70.9}{2.2} \\
 & GEPA & \dvalp{\underline{79.4}}{3.5} & \dvalp{42.5}{6.4} & \dvalp{45.3}{23.2} & \dvalp{83.7}{12.3} & \dvalp{27.2}{12.5} & \dvalp{81.3}{8.2} \\
 & \methodcell{\textbf{\ourmethod{}}} & \methodcell{\dvalp{\textbf{80.2}}{4.3}} & \methodcell{\dvalp{\textbf{47.5}}{11.4}} & \methodcell{\dvalp{\textbf{48.8}}{26.7}} & \methodcell{\dvalp{\textbf{90.9}}{19.5}} & \methodcell{\dvalp{\textbf{32.8}}{18.1}} & \methodcell{\dvalp{\textbf{85.8}}{12.7}} \\
\cmidrule(lr){1-8}
\multirow{7}{*}{GPT--5.4-nano} & No skill & 55.8 & 23.5 & 16.3 & 30.8 & 23.2 & 34.3 \\
 & Human skill & \dvalp{66.1}{10.3} & \dvalp{\underline{41.8}}{18.3} & \dvalp{\underline{46.5}}{30.2} & \dvalp{73.5}{42.7} & \dvaln{16.8}{6.4} & \dvaln{29.9}{4.4} \\
 & LLM skill & \dvalp{62.4}{6.6} & \dvalp{33.6}{10.1} & \dvaln{12.8}{3.5} & \dvalp{71.7}{40.9} & \dvaln{20.0}{3.2} & \dvalp{65.7}{31.4} \\
 & Trace2Skill & \dvalp{69.9}{14.1} & \dvalp{35.4}{11.9} & \dvalp{16.3}{0.0} & \dvalp{\underline{77.3}}{46.5} & \dvalp{\underline{25.6}}{2.4} & \dvalp{58.2}{23.9} \\
 & TextGrad & \dvalp{\underline{73.4}}{17.6} & \dvalp{32.5}{9.0} & \dvalp{38.7}{22.4} & \dvalp{68.3}{37.5} & \dvaln{20.8}{2.4} & \dvalp{42.5}{8.2} \\
 & GEPA & \dvalp{73.2}{17.4} & \dvalp{37.2}{13.7} & \dvalp{45.0}{28.7} & \dvalp{71.2}{40.4} & \dvalp{24.8}{1.6} & \dvalp{\underline{68.7}}{34.4} \\
 & \methodcell{\textbf{\ourmethod{}}} & \methodcell{\dvalp{\textbf{74.8}}{19.0}} & \methodcell{\dvalp{\textbf{42.5}}{19.0}} & \methodcell{\dvalp{\textbf{50.0}}{33.7}} & \methodcell{\dvalp{\textbf{80.2}}{49.4}} & \methodcell{\dvalp{\textbf{27.2}}{4.0}} & \methodcell{\dvalp{\textbf{69.4}}{35.1}} \\
\cmidrule(lr){1-8}
\multirow{7}{*}{GPT--5.2} & No skill & 71.9 & 38.2 & 34.9 & 73.1 & 20.8 & 68.7 \\
 & Human skill & \dvalp{75.8}{3.9} & \dvalp{48.2}{10.0} & \dvalp{\underline{55.2}}{20.3} & \dvalp{89.0}{15.9} & \dvalp{23.2}{2.4} & \dvaln{56.7}{12.0} \\
 & LLM skill & \dvalp{73.7}{1.8} & \dvalp{38.6}{0.4} & \dvaln{14.0}{20.9} & \dvalp{87.7}{14.6} & \dvalp{25.6}{4.8} & \dvalp{68.7}{0.0} \\
 & Trace2Skill & \dvalp{79.2}{7.3} & \dvalp{43.9}{5.7} & \dvalp{43.0}{8.1} & \dvalp{87.7}{14.6} & \dvalp{\underline{28.8}}{8.0} & \dvalp{76.9}{8.2} \\
 & TextGrad & \dvalp{82.6}{10.7} & \dvaln{32.1}{6.1} & \dvalp{51.1}{16.2} & \dvalp{87.2}{14.1} & \dvalp{22.4}{1.6} & \dvalp{76.1}{7.4} \\
 & GEPA & \dvalp{\underline{82.7}}{10.8} & \dvalp{\underline{54.3}}{16.1} & \dvalp{53.4}{18.5} & \dvalp{\underline{89.5}}{16.4} & \dvalp{\underline{28.8}}{8.0} & \dvalp{\underline{82.8}}{14.1} \\
 & \methodcell{\textbf{\ourmethod{}}} & \methodcell{\dvalp{\textbf{83.1}}{11.2}} & \methodcell{\dvalp{\textbf{57.1}}{18.9}} & \methodcell{\dvalp{\textbf{56.4}}{21.5}} & \methodcell{\dvalp{\textbf{89.6}}{16.5}} & \methodcell{\dvalp{\textbf{36.0}}{15.2}} & \methodcell{\dvalp{\textbf{85.1}}{16.4}} \\
\cmidrule(lr){1-8}
\multirow{7}{*}{Qwen3.5--4B} & No skill & 68.1 & 9.3 & 14.5 & 86.9 & 22.4 & 30.6 \\
 & Human skill & \dvaln{66.3}{1.8} & \dvalp{16.4}{7.1} & \dvalp{\underline{22.7}}{8.2} & \dvalp{87.8}{0.9} & \dvaln{18.4}{4.0} & \dvaln{28.4}{2.2} \\
 & LLM skill & \dvaln{65.0}{3.1} & \dvalp{10.4}{1.1} & \dvalp{20.9}{6.4} & \dvalp{\underline{88.0}}{1.1} & \dvalp{\underline{28.8}}{6.4} & \dvalp{55.2}{24.6} \\
 & Trace2Skill & \dvalp{68.5}{0.4} & \dvalp{\underline{19.3}}{10.0} & \dvalp{16.3}{1.8} & \dvalp{\underline{88.0}}{1.1} & \dvalp{27.2}{4.8} & \dvalp{\underline{64.9}}{34.3} \\
 & TextGrad & \dvaln{60.7}{7.4} & \dvalp{13.9}{4.6} & \dvalp{20.9}{6.4} & \dvaln{85.6}{1.3} & \dvaln{10.6}{11.8} & \dvalp{53.7}{23.1} \\
 & GEPA & \dvalp{\underline{68.6}}{0.5} & \dvalp{16.9}{7.6} & \dvalp{\underline{22.7}}{8.2} & \dvaln{85.1}{1.8} & \dvalp{\underline{28.8}}{6.4} & \dvalp{60.4}{29.8} \\
 & \methodcell{\textbf{\ourmethod{}}} & \methodcell{\dvalp{\textbf{71.2}}{3.1}} & \methodcell{\dvalp{\textbf{23.9}}{14.6}} & \methodcell{\dvalp{\textbf{29.7}}{15.2}} & \methodcell{\dvalp{\textbf{89.0}}{2.1}} & \methodcell{\dvalp{\textbf{52.0}}{29.6}} & \methodcell{\dvalp{\textbf{81.3}}{50.7}} \\
\cmidrule(lr){1-8}
\multirow{7}{*}{Qwen3.6--35B-A3B} & No skill & 72.7 & 38.2 & 45.9 & 87.6 & \underline{31.2} & 59.7 \\
 & Human skill & \dvalp{74.1}{1.4} & \dvalp{44.3}{6.1} & \dvaln{41.9}{4.0} & \dvalp{88.4}{0.8} & \dvaln{29.6}{1.6} & \dvaln{44.8}{14.9} \\
 & LLM skill & \dvaln{72.6}{0.1} & \dvalp{42.9}{4.7} & \dvalp{\underline{46.5}}{0.6} & \dvalp{88.4}{0.8} & \dvaln{24.8}{6.4} & \dvalp{60.4}{0.7} \\
 & Trace2Skill & \dvalp{75.4}{2.7} & \dvaln{33.2}{5.0} & \dvaln{32.0}{13.9} & \dvalp{\underline{90.4}}{2.8} & \dvaln{29.6}{1.6} & \dvalp{70.9}{11.2} \\
 & TextGrad & \dvalp{\underline{76.4}}{3.7} & \dvaln{22.9}{15.3} & \dvaln{33.7}{12.2} & \dvaln{84.5}{3.1} & \dvaln{7.2}{24.0} & \dvalp{67.9}{8.2} \\
 & GEPA & \dvalp{75.8}{3.1} & \dvalp{\underline{45.4}}{7.2} & \dvaln{43.6}{2.3} & \dvalp{88.0}{0.4} & \dvalp{\underline{31.2}}{0.0} & \dvalp{\underline{73.9}}{14.2} \\
 & \methodcell{\textbf{\ourmethod{}}} & \methodcell{\dvalp{\textbf{80.3}}{7.6}} & \methodcell{\dvalp{\textbf{47.5}}{9.3}} & \methodcell{\dvalp{\textbf{47.1}}{1.2}} & \methodcell{\dvalp{\textbf{91.4}}{3.8}} & \methodcell{\dvalp{\textbf{41.6}}{10.4}} & \methodcell{\dvalp{\textbf{82.1}}{22.4}} \\
\midrule
\harnessrow{Codex harness} \\
\multirow{5}{*}{GPT--5.5} & No skill & 81.8 & 27.5 & 38.3 & 87.2 & 35.2 & \na \\
 & Human skill & \dvalp{\underline{84.1}}{2.3} & \dvalp{50.7}{23.2} & \dvalp{40.0}{1.7} & \dvalp{88.8}{1.6} & \dvalp{48.8}{13.6} & \na \\
 & LLM skill & \dvalp{83.4}{1.6} & \dvaln{25.0}{2.5} & \dvaln{34.3}{4.0} & \dvalp{\underline{89.8}}{2.6} & \dvalp{45.6}{10.4} & \na \\
 & EvoSkill & \dvaln{61.4}{20.4} & \dvalp{\underline{67.5}}{40.0} & \dvalp{\underline{42.4}}{4.1} & \dvalp{89.3}{2.1} & \dvalp{\underline{63.2}}{28.0} & \na \\
 & \methodcell{\textbf{\ourmethod{}}} & \methodcell{\dvalp{\textbf{87.3}}{5.5}} & \methodcell{\dvalp{\textbf{85.0}}{57.5}} & \methodcell{\dvalp{\textbf{51.1}}{12.8}} & \methodcell{\dvalp{\textbf{92.2}}{5.0}} & \methodcell{\dvalp{\textbf{78.4}}{43.2}} & \methodcell{\na} \\
\midrule
\harnessrow{Claude Code harness} \\
\multirow{5}{*}{GPT--5.5} & No skill & 81.9 & 22.1 & 57.6 & 86.6 & 40.8 & \na \\
 & Human skill & \dvalp{83.7}{1.8} & \dvalp{37.1}{15.0} & \dvalp{66.3}{8.7} & \dvalp{88.0}{1.4} & \dvalp{44.0}{3.2} & \na \\
 & LLM skill & \dvalp{82.4}{0.5} & \dvalp{37.1}{15.0} & \dvaln{56.4}{1.2} & \dvalp{\underline{89.6}}{3.0} & \dvalp{44.8}{4.0} & \na \\
 & EvoSkill & \dvalp{\underline{84.0}}{2.1} & \dvalp{\underline{75.0}}{52.9} & \dvalp{\underline{70.3}}{12.7} & \dvalp{87.2}{0.6} & \dvalp{\underline{52.0}}{11.2} & \na \\
 & \methodcell{\textbf{\ourmethod{}}} & \methodcell{\dvalp{\textbf{85.9}}{4.0}} & \methodcell{\dvalp{\textbf{80.4}}{58.3}} & \methodcell{\dvalp{\textbf{71.5}}{13.9}} & \methodcell{\dvalp{\textbf{90.1}}{3.5}} & \methodcell{\dvalp{\textbf{56.5}}{15.7}} & \methodcell{\na} \\
\bottomrule
\end{tabular}
\caption{Main results on held-out test splits. Scores are percentages; within each model--harness block, bold marks the best measured entry and underlining marks the second-best entry for each benchmark. Blue cells denote \ourmethod{}, and small green/red subscripts show the absolute change relative to the \emph{No skill} row of the same model in the same harness. We omit ALFWorld under Codex and Claude Code harnesses because ALFWorld requires persistent embodied-environment interaction. \ourmethod{} is the best-or-tied entry on every measured cell of the table, with positive gains over the no-skill baseline throughout.}
\label{tab:main_results_by_harness}
\end{table*}

\subsection{Epoch-Wise Slow/Meta Update}
\label{sec:method_slow_meta}

Fast updates learn from the current batch; the epoch-wise slow/meta update learns from adjacent epochs. At the end of an epoch, \ourmethod{} samples the same training items under the previous epoch's skill and the current skill, then groups them into improvements, regressions, persistent failures, and stable successes. The optimizer model writes a concise longitudinal guidance block into a protected slow-update field, and this candidate is still passed through the validation gate. Thus slow update captures durable domain lessons while preserving the same safety check as step-level edits.

The meta skill is optimizer-side only. It summarizes which edit patterns helped, which were rejected, and which failures persisted across epochs. This meta guidance is prepended to future optimizer prompts for reflection, merging, and ranking, but it is not shipped with the target model. The advantage is separation of concerns: the deployed skill remains compact and portable, while training benefits from a richer record of the editing process.

\subsection{Harness-Agnostic Deployment}
\label{sec:method_harness}

\ourmethod{} is harness-agnostic through a lightweight adapter interface, matching the broader trend toward agents embedded in tool-use and software-execution environments \citep{yao2022react,schick2023toolformer,yang2024sweagent}. An adapter constructs train/evaluation batches, injects the current skill into the agent context, runs the native harness, and returns scored trajectories. The same optimizer therefore works for direct QA, spreadsheet execution, document reasoning, multimodal QA, embodied environments, and Codex-style or Claude Code-style execution loops. This is the main practical advantage of treating skills as the adaptation layer: a stronger optimizer model can train a reusable skill artifact offline, and the resulting \texttt{best\_skill.md} can then be deployed or tested across target models, harnesses, and nearby benchmarks without changing model weights.

\begin{table*}[t]
\centering
\scriptsize
\setlength{\tabcolsep}{1.2pt}
\renewcommand{\arraystretch}{0.74}
\resizebox{\textwidth}{!}{%
\begin{minipage}{1.12\textwidth}
\begin{minipage}[c]{0.33\textwidth}
\centering
\textbf{(a) Training set size}\\[-1pt]
\begin{tabular}{@{}lccc@{}}
\toprule
\ablationbenchmarkheader
\midrule
1 example & 81.0 & 47.5 & 59.1 \\
20\% train & 84.1 & 69.0 & 65.9 \\
40\% train & \textbf{86.1} & 73.5 & 64.8 \\
80\% train & \textbf{86.1} & 77.6 & 67.0 \\
100\% train & 84.1 & \textbf{78.0} & \textbf{70.5} \\
\bottomrule
\end{tabular}
\end{minipage}
\hfill
\begin{minipage}[c]{0.33\textwidth}
\centering
\textbf{(b) Mini-batchsize}\\[-1pt]
\begin{tabular}{@{}lccc@{}}
\toprule
\ablationbenchmarkheader
\midrule
1 & 85.9 & 75.4 & 60.5 \\
2 & 86.3 & 77.1 & 54.8 \\
4 & 86.9 & 75.4 & \textbf{64.5} \\
8 & \textbf{87.1} & 77.5 & 61.3 \\
16 & 87.0 & \textbf{77.9} & 61.3 \\
32 & 86.9 & 77.5 & 58.9 \\
\bottomrule
\end{tabular}
\end{minipage}
\hfill
\begin{minipage}[c]{0.33\textwidth}
\centering
\textbf{(c) Batchsize}\\[-1pt]
\begin{tabular}{@{}lccc@{}}
\toprule
\ablationbenchmarkheader
\midrule
8 & 85.1 & 76.8 & 58.1 \\
24 & 86.4 & 77.1 & \textbf{62.9} \\
40 & 87.1 & \textbf{77.5} & 61.3 \\
56 & 86.5 & 76.8 & 56.5 \\
full epoch & \textbf{87.2} & 75.0 & 53.2 \\
\bottomrule
\end{tabular}
\end{minipage}

\vspace{2pt}

\begin{minipage}[c]{0.33\textwidth}
\centering
\textbf{(d) Learning rate}\\[-1pt]
\begin{tabular}{@{}lccc@{}}
\toprule
\ablationbenchmarkheader
\midrule
lr=1 & 85.5 & 77.5 & 62.1 \\
lr=2 & 86.7 & 77.5 & 60.5 \\
lr=4 & 86.5 & \textbf{78.2} & 56.5 \\
lr=8 & \textbf{87.0} & 73.6 & \textbf{66.9} \\
lr=16 & 86.8 & \textbf{78.2} & 65.3 \\
\bottomrule
\end{tabular}
\end{minipage}
\hfill
\begin{minipage}[c]{0.33\textwidth}
\centering
\textbf{(e) Learning-rate scheduler}\\[-1pt]
\begin{tabular}{@{}lccc@{}}
\toprule
\ablationbenchmarkheader
\midrule
constant & \textbf{87.3} & \textbf{80.7} & 62.1 \\
cosine & 87.1 & 77.5 & 61.3 \\
linear & 87.2 & 72.9 & \textbf{62.9} \\
\bottomrule
\end{tabular}
\end{minipage}
\hfill
\begin{minipage}[c]{0.33\textwidth}
\centering
\textbf{(f) Slow-update samples}\\[-1pt]
\begin{tabular}{@{}lccc@{}}
\toprule
\ablationbenchmarkheader
\midrule
5 & 86.8 & 76.4 & 64.5 \\
10 & 86.4 & 74.3 & \textbf{65.3} \\
20 & \textbf{87.1} & \textbf{77.5} & 61.3 \\
40 & 86.9 & 75.4 & 54.8 \\
\bottomrule
\end{tabular}
\end{minipage}
\end{minipage}%
}
\caption{Hyperparameter analysis for the text optimizer. Each panel changes one scalar or scheduling factor from the default setting unless noted. Panel (a) fixes the split to $4{:}1{:}5$ train/selection/test; the 1-example, 20\%, 40\%, and 80\% rows use subsets of the training partition, and the 100\% row reuses the completed $4{:}1{:}5$ split-ratio run. Panel (b) sweeps the reflection mini-batchsize $B_m$; panel (c) sweeps the rollout batchsize $B$.}
\label{tab:ablation_sweeps}
\end{table*}

\begin{table*}[t]
\centering
\scriptsize
\setlength{\tabcolsep}{4pt}
\renewcommand{\arraystretch}{0.82}
\resizebox{0.92\textwidth}{!}{%
\begin{tabular}{l l c c c}
\toprule
\textbf{Component} & \textbf{Setting} & \textbf{SearchQA} & \textbf{SpreadsheetBench} & \textbf{LiveMath} \\
\midrule
\rowcolor{blue!8}Learning-rate form & lr=4 (default) & \textbf{87.1} & \textbf{77.5} & \textbf{61.3} \\
 & dynamic lr & 85.8 & 71.8 & 54.0 \\
 & without lr & 84.6 & 75.7 & 57.3\\
\cmidrule(lr){1-5}
\rowcolor{blue!8}Rejected buffer & with rejected buffer & \textbf{87.1} & \textbf{77.5} & \textbf{61.3} \\
 & without rejected buffer & 85.5 & 72.9 & 58.9 \\
\cmidrule(lr){1-5}
\rowcolor{blue!8}Slow/meta update & meta skill and slow update & \textbf{87.1} & \textbf{77.5} & \textbf{61.3} \\
 & without meta skill & 85.1 & 75.7 & 58.1 \\
 & without meta skill and slow update & 86.3 & 55.0 & 59.7 \\
\bottomrule
\end{tabular}%
}
\caption{Component ablations for learning-rate form, rejected buffer, and epoch-wise slow/meta update. Light-blue rows mark the default setting within each component group; the learning-rate group uses the default lr=4 setting. Bold values mark the best measured result within that group and benchmark. The without-rejected-buffer row uses the matched no-buffer ablation setting.}
\label{tab:component_ablation}
\end{table*}

\begin{table*}[t]
\centering
\scriptsize
\setlength{\tabcolsep}{6pt}
\renewcommand{\arraystretch}{0.9}
\resizebox{0.95\textwidth}{!}{%
\begin{tabular}{l l l c c c}
\toprule
\rowcolor{black!7}\multicolumn{6}{l}{\textbf{(a) Cross-model transfer}} \\
\midrule
\textbf{Source model} & \textbf{Target model} & \textbf{Benchmark} & \textbf{Baseline} & \textbf{Direct} & \textbf{Transferred} \\
\midrule
\multirow{3}{*}{GPT--5.4} & GPT--5.4      & \multirow{3}{*}{SpreadsheetBench} & 41.4 & 62.5 & \na \\
                          & GPT--5.4-mini &                                    & 36.1 & 47.5 & \dvalp{45.5}{9.4}  \\
                          & GPT--5.4-nano &                                    & 23.5 & 42.5 & \dvalp{26.5}{3.0}  \\
\cmidrule(lr){1-6}
\multirow{3}{*}{GPT--5.4} & GPT--5.4      & \multirow{3}{*}{LiveMath}         & 36.8 & 44.0 & \na \\
                          & GPT--5.4-mini &                                    & 14.7 & 32.8 & \dvalp{19.2}{4.5}  \\
                          & GPT--5.4-nano &                                    & 23.2 & 27.2 & \dvalp{28.8}{5.6}  \\
\midrule
\rowcolor{black!7}\multicolumn{6}{l}{\textbf{(b) Cross-harness transfer}} \\
\midrule
\textbf{Source harness} & \textbf{Target harness} & \textbf{Benchmark} & \textbf{Baseline} & \textbf{Direct} & \textbf{Transferred} \\
\midrule
Codex                       & Claude Code & \multirow{2}{*}{LiveMath}         & 40.8 & 56.5 & \dvalp{42.4}{1.6}  \\
Claude Code                 & Codex       &                                    & 35.2 & 78.4 & \dvalp{48.0}{12.8} \\
Codex                       & Claude Code & \multirow{2}{*}{SpreadsheetBench} & 22.1 & 80.4 & \dvalp{81.8}{59.7} \\
Claude Code                 & Codex       &                                    & 27.5 & 85.0 & \dvalp{71.1}{43.6} \\
\midrule
\rowcolor{black!7}\multicolumn{6}{l}{\textbf{(c) Cross-benchmark transfer}} \\
\midrule
\textbf{Source benchmark} & \textbf{Target benchmark} & \textbf{Model} & \textbf{Baseline} & \textbf{Direct} & \textbf{Transferred} \\
\midrule
\multirow{3}{*}{OlympiadBench} & \multirow{3}{*}{Omni-MATH} & GPT--5.4      & 56.6 & \na & \dvalp{60.3}{3.7} \\
                                & & GPT--5.4-mini & 34.8 & \na & \dvalp{36.6}{1.8} \\
                                & & GPT--5.4-nano & 38.8 & \na & \dvalp{40.1}{1.3} \\
\bottomrule
\end{tabular}%
}
\caption{Transfer of optimized skills across three axes. \textbf{(a)} \emph{Cross-model}: a skill optimized for the source model is deployed on the target model. \textbf{(b)} \emph{Cross-harness}: a skill trained inside the source harness is evaluated inside the target harness, all on GPT--5.5. \textbf{(c)} \emph{Cross-benchmark}: the source benchmark skill is evaluated on the target benchmark across three target models. \textbf{Baseline} is the target's no-skill score, \textbf{Direct} is the in-domain \ourmethod{} score, and \textbf{Transferred} applies the source skill without further optimization. Subscripts show the change over the target baseline. The GPT--5.4{$\to$}GPT--5.4 transferred cells in (a) are marked \na{} because source and target match (i.e.\ no transfer occurs); we still report the GPT--5.4 baseline and in-domain \ourmethod{} score (taken from Table~\ref{tab:main_results_by_harness}) in those rows for reference. Every remaining row in (a)--(c) is a positive transfer (no row falls below the target's no-skill baseline).}
\label{tab:transfer_all}
\label{tab:transfer_cross_model}
\label{tab:transfer_cross_harness}
\label{tab:transfer_cross_benchmark}
\end{table*}

\providecommand{\ourmethod}{\textsc{SkillOpt}}
\providecommand{\na}{--}
\providecommand{\harnessrow}[1]{\rowcolor{black!7}\multicolumn{8}{l}{\textbf{#1}}}
\providecommand{\methodcell}[1]{\cellcolor{blue!8}#1}
\providecommand{\ablationbenchmarkheader}{\textbf{Setting} & \textbf{SearchQA} & \textbf{Spreadsheet} & \textbf{LiveMath} \\}

\section{Experiments}
\label{sec:experiments}

We evaluate \ourmethod{} as a text-space optimizer for frozen agents: the target model executes each task with the current skill, while an offline optimizer edits that skill from rollout evidence. The experiments answer four questions. (i) Do optimized skills improve over no-skill, human-skill, one-shot LLM-skill, prompt-optimization (TextGrad, GEPA), and skill-evolution (Trace2Skill, EvoSkill) baselines? (ii) Does the same loop work across direct chat, Codex, and Claude Code harnesses, and across seven target models from frontier-scale GPT to small Qwen? (iii) Which optimizer controls matter? (iv) What do the learned skills look like, and at what cost?

\paragraph{Setting.}
We report each benchmark's native hard score or exact-match accuracy on held-out test splits across SearchQA~\citep{dunn2017searchqa}, SpreadsheetBench~\citep{spreadsheetbench}, OfficeQA~\citep{opsahl2026officeqa}, DocVQA~\citep{mathew2021docvqa}, LiveMathematicianBench~\citep{he2026livemathematicianbenchlivebenchmarkmathematicianlevel} (abbreviated LiveMath in tables), and ALFWorld~\citep{alfworld}, using two model families: GPT~\citep{openai2026gpt54} and Qwen~\citep{qwen3.5,qwen36_35b_a3b}. The benchmark suite is intentionally diverse---it spans single-round QA (SearchQA, DocVQA, LiveMathematicianBench MCQ), multi-turn tool loops with up to $24$ tool calls (OfficeQA), multi-round codegen with up to $30$ turns and a real \texttt{openpyxl}/\texttt{pandas} runtime (SpreadsheetBench, default \texttt{mode=multi}), and persistent embodied interaction with up to $50$ steps per episode (ALFWorld). Dataset-backed runs use deterministic train/selection/test splits derived from the same dataset seed ($\mathtt{split\_seed=42}$); the selection split is used \emph{only} to accept or reject candidate skill edits, and all reported scores are computed on the disjoint held-out test split. The reported numbers thus measure generalization, not validation-set fit.

\paragraph{Default optimizer hyperparameters.}
Unless noted, \ourmethod{} uses four epochs, rollout batch size $40$ per step, reflection minibatch size $8$ (with $16$ analyst workers running reflections in parallel and a merge batch size of $8$), textual learning rate $L_t=4$ with cosine decay (floor $L_t=2$, configurable schedules: constant, linear, cosine, autonomous), held-out validation gating (strictly greater than the current selection score---ties are rejected), slow update with $20$ sampled tasks per epoch comparing previous-epoch and current-epoch skill, an optimizer-side meta skill that summarizes accepted/rejected patterns into teacher-only guidance, the \texttt{patch} edit mode (the alternative is \texttt{rewrite\_from\_suggestions}), and an optional rejected-edit buffer of recent failed proposals. Teacher reflection is allowed up to three refinement rounds per minibatch. Both teacher and student calls default to a \texttt{medium} reasoning effort. For benchmarks with tightly bounded training pools (LiveMathematicianBench: $35$ training items per epoch with rollout batch $200$; ALFWorld: $39$ training tasks with $140$ selection and $134$ test environments), per-benchmark configs scale the batch sizes accordingly while keeping the same gate, scheduler, and slow/meta-update machinery. Additional benchmark, baseline, and optimizer-protocol details are in Appendix~\ref{app:experimental_setting}.

\paragraph{Harnesses.}
Direct chat invokes the target model through a single chat completion call with the skill prepended to the system prompt. The Codex harness drives the target through the \texttt{codex} CLI in a workspace-write sandbox~\citep{openai2025codex}; \ourmethod{} renders the current skill to a per-task \texttt{SKILL.md} alongside task files and reads back a compact execution trace (\texttt{codex\_trace\_summary.txt}) that is included in the teacher reflection context, so the optimizer learns from \emph{what the agent actually did}, not just its final answer. The Claude Code harness mirrors the same workspace contract through the \texttt{claude} CLI~\citep{anthropic2025claudecode}. All three modes consume the same \texttt{best\_skill.md} file format, which is what enables the cross-harness transfer experiments in Section~\ref{sec:analysis_transfer}.

\paragraph{Baselines.}
We compare against seven baselines that span the no-adaptation, hand-written, one-shot, and learning families: \emph{no skill} (frozen target model run with the benchmark's default system prompt), \emph{human skill} (an expert-written skill document curated per benchmark), \emph{one-shot LLM skill} (a single skill generated from a high-level task description by GPT--5.5 and never updated), \emph{Trace2Skill}~\citep{ni2026trace2skill} (trajectory-level skill distillation), \emph{TextGrad}~\citep{yuksekgonul2024textgrad} (gradient-style natural-language prompt optimization), \emph{GEPA}~\citep{agrawal2025gepa} (Pareto reflective prompt evolution), and the harness-side competitor \emph{EvoSkill}~\citep{alzubi2026evoskill} (skill-folder evolution under failure analysis). All baselines use the same target model, the same held-out test split, and the same scorer for every benchmark, so the comparison isolates the choice of adaptation procedure rather than secondary factors such as prompt template or scoring pipeline.

\subsection{Main Results}
\label{sec:main_results}

Table~\ref{tab:main_results_by_harness} is the main result matrix. Counting every (target model, benchmark, harness) cell as one comparison and the strongest of the no-skill, human-skill, LLM-skill, Trace2Skill, TextGrad, GEPA, and EvoSkill baselines as the per-cell competition, \ourmethod{} wins or matches the best measured result on \textbf{$52$ of $52$ evaluated cells}. This dominance is uniform across model scales: \ourmethod{} is best on every benchmark for GPT--5.5, GPT--5.4, GPT--5.4-mini, GPT--5.4-nano, GPT--5.2, Qwen3.5--4B, and Qwen3.6--35B-A3B in direct chat, and for GPT--5.5 under both Codex and Claude Code harnesses.

The size of the gains is also unusually large for a no-weight-update method. On GPT--5.5 direct chat, the six-benchmark average rises from $58.8$ (no skill) to $82.3$ (\ourmethod{}), a $+23.5$ point absolute improvement, while the best per-cell baseline averages only $76.9$, leaving \ourmethod{} $+5.4$ points clear of an oracle baseline that picks the best of six competing methods per cell. Per-benchmark deltas over no skill range from $+9.6$ on SearchQA, where the no-skill model is already near ceiling, to $+38.9$ on SpreadsheetBench and $+39.0$ on OfficeQA, where strict procedural and answer-format requirements expose the limits of zero-shot frontier models. Procedural benchmarks see the largest improvements: SpreadsheetBench $41.8{\to}80.7$, OfficeQA $33.1{\to}72.1$, and LiveMathematicianBench $37.6{\to}66.9$ on GPT--5.5; SpreadsheetBench $9.3{\to}23.9$ ($\times 2.6$) on Qwen3.5--4B; and ALFWorld $34.3{\to}69.4$ ($\times 2.0$) on GPT--5.4-nano.

The improvement is not specific to frontier-scale targets. Averaged over the six benchmarks, \ourmethod{} lifts GPT--5.4 by $+12.7$ points, GPT--5.4-mini by $+15.4$, GPT--5.4-nano by $+26.7$, GPT--5.2 by $+16.6$, Qwen3.5--4B by $+19.2$, and Qwen3.6--35B-A3B by $+9.1$, for an average improvement of approximately $+17.6$ points per model. Small and weak target models benefit the most in relative terms (e.g.\ GPT--5.4-nano nearly doubles on DocVQA and triples on ALFWorld), which is consistent with the view that a compact skill artifact can supply procedural knowledge that small models do not yet hold in weights.

The same optimization interface is also effective under tool-backed execution. On the Codex harness, \ourmethod{} is best on all five evaluated benchmarks for GPT--5.5, with average gain $+24.8$ points over no skill and $+14.0$ over the next-best baseline (EvoSkill). On the Claude Code harness, it is best on all five benchmarks for GPT--5.5, with average gain $+19.1$ over no skill and $+3.2$ over EvoSkill, while EvoSkill itself already lifts the five-benchmark average from $57.8$ to $73.7$. The two ALFWorld cells under harness rows are left blank because ALFWorld requires persistent embodied-environment interaction that is not represented in the standard Codex / Claude Code adapters; we therefore report harness results on search, spreadsheets, document QA, multimodal QA, and math.

Taken together, the table supports a strong empirical claim: across direct chat and two tool-execution harnesses, across seven target models, and on procedural and factual benchmarks alike, optimizing a single compact skill artifact under bounded text-space training is the strongest no-weight-update adaptation strategy among the baselines we consider. The main gains come from feedback-driven skill editing rather than from a better one-shot prompt: human and LLM skills can help when prior instructions happen to match the benchmark, but they cannot correct failures after observing rollouts; Trace2Skill mines trajectory lessons without a held-out gate; TextGrad and GEPA optimize prompts but not a persistent skill artifact; and EvoSkill, the strongest harness-side competitor, lacks both bounded textual learning rates and rejected-edit memory. These comparisons support the central design choice---keep the target model, harness, and evaluator fixed, and optimize only the reusable skill artifact.

\paragraph{Alternative explanations.}
The per-cell baselines clarify what drives the gains. The effect is not simply prompt length: human skills are already $145$--$516$ tokens long and often exceed the one-shot LLM skill, yet they are beaten in every direct-chat model row while the learned artifacts remain compact (Table~\ref{tab:skill_cost_case}). It is also not only optimizer capacity: \ourmethod{} leads every baseline even for GPT--5.4-nano, and the optimizer-strength analysis in Table~\ref{tab:teacher_model_ablation} shows that a target-matched optimizer recovers much of the gain. Finally, the harness results show that the method is not just exploiting one skill format: EvoSkill already improves the Codex SpreadsheetBench cell from $27.5$ to $67.5$, but \ourmethod{} adds another $+17.5$ points ($67.5{\to}85.0$). The gains are largest on procedural benchmarks, where reusable rules about tool use and output formatting matter most, but they also appear on factual and multimodal benchmarks.

\paragraph{Headline numbers in one place.}
For convenience, the headline aggregates over Table~\ref{tab:main_results_by_harness} are:
(i) \textbf{$52/52$} cells best or tied-best;
(ii) average per-model improvement $\approx +17.6$ points across the seven direct-chat target models;
(iii) average GPT--5.5 improvement of $+23.5$ (direct chat), $+24.8$ (Codex), $+19.1$ (Claude Code) over no skill;
(iv) GPT--5.5 oracle-baseline gap of $+5.4$ points (direct chat) computed as the difference between \ourmethod{}'s six-benchmark average ($82.3$) and an oracle that picks the best of six competing methods \emph{per cell} ($76.9$).
The remainder of this section unpacks why these gains appear (Section~\ref{sec:optimizer_ablation}), how stable and transferable they are (Section~\ref{sec:analysis_transfer}), and what the learned artifact looks like (Section~\ref{sec:learned_skills}).

\subsection{Ablations}
\label{sec:optimizer_ablation}

\begin{figure*}[t]
\centering
\includegraphics[width=0.92\textwidth]{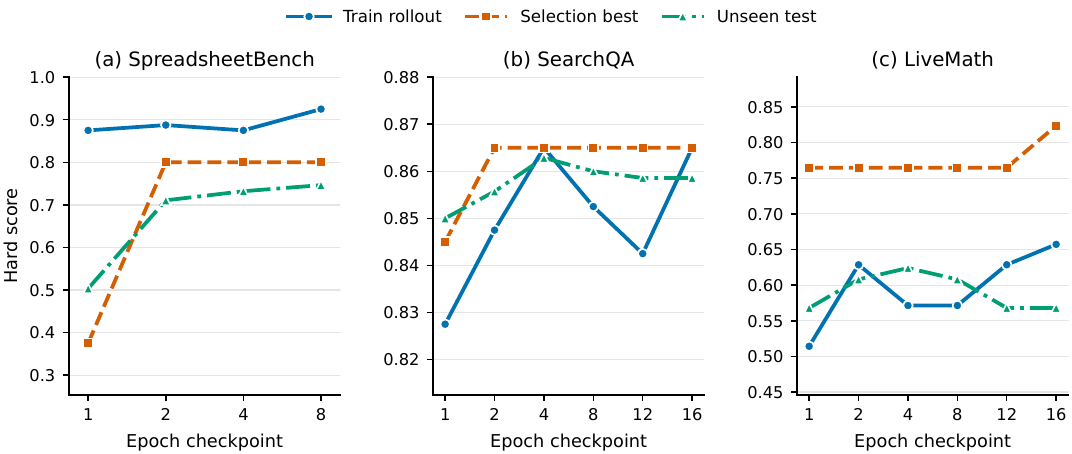}
\caption{Performance trends across epoch checkpoints on three benchmarks: (a) SpreadsheetBench, (b) SearchQA, and (c) LiveMath. For each checkpoint, we report the training rollout score, the selection-best score on the validation set, and the final performance on the unseen test set. The results show how skill quality evolves during optimization and whether the checkpoint preferred by validation selection aligns with the checkpoint that yields the best generalization to the test set.}
\label{fig:epoch_accuracy_curves}
\end{figure*}

Table~\ref{tab:ablation_sweeps}, Figure~\ref{fig:epoch_accuracy_curves}, and Table~\ref{tab:component_ablation} test the design choices in the optimizer using GPT--5.5 as both the target and the optimizer. The overall message is that \ourmethod{} benefits from sufficient evidence, a bounded textual learning rate, rejected-edit feedback, and epoch-wise slow/meta update. SearchQA has limited headroom and is therefore stable across many settings (most cells fluctuate inside a $\pm1.5$ point band), while SpreadsheetBench and LiveMathematicianBench expose the trade-off between learning useful procedures and over-editing the skill.

\paragraph{Evidence and batch sizes (panels a, b, c).}
Panel (a) shows that procedural benchmarks reward more training evidence: SpreadsheetBench climbs $47.5{\to}78.0$ and LiveMathematicianBench climbs $59.1{\to}70.5$ as the optimizer sees $1{\to}100\%$ of the training partition, while SearchQA saturates at roughly $84{-}86$ after $20\%$ already. Panel (b) shows the same robustness in the other direction: varying the reflection mini-batchsize from $1$ to $32$ keeps SearchQA inside $85.9{-}87.1$ and SpreadsheetBench inside $75.4{-}77.9$, with the default $B_m{=}8$ at or near the top on all three benchmarks. Panel (c) is equally flat in the rollout-batchsize dimension---moving from $B{=}8$ to a full epoch keeps SearchQA inside $85.1{-}87.2$ and SpreadsheetBench inside $75.0{-}77.5$. Together this means the headline gains are not the product of a fragile prompt-search batch size, but a genuine effect of having enough scored evidence per update.

\paragraph{Textual learning rate and schedule (panels d, e).}
Panels (d) and (e) directly compare bounded textual learning to looser settings. Sweeping $L_t \in \{1,2,4,8,16\}$ shows that small or moderate edit budgets are competitive throughout: $L_t{=}4$ achieves $86.5/78.2/56.5$, the highest LiveMath score belongs to $L_t{=}8$ at $66.9$, and the lowest score across all five settings is still only $85.5$ on SearchQA. Panel (e) confirms this on the schedule axis: constant decay scores $87.3/80.7/62.1$, cosine $87.1/77.5/61.3$, and linear $87.2/72.9/62.9$, so the bounded-update story does not depend on a single specific scheduler. The qualitative claim is what matters: any moderate, bounded edit budget already beats baselines that rewrite the skill without a budget (Table~\ref{tab:component_ablation}, ``without lr'' row, $84.6/75.7/57.3$).

\paragraph{Epoch-wise slow/meta update (panel f, Table~\ref{tab:component_ablation}, Figure~\ref{fig:epoch_accuracy_curves}).}
The slow/meta update supplies longer-horizon guidance beyond the current rollout batch. Slow-update sampling (panel f) places the default at $20$ examples per epoch (87.1, 77.5, and 61.3), with $5$, $10$, and $40$ each within $\pm 2.7$ points. In the matched default component row, removing the rejected-edit buffer lowers scores by 1.6, 4.6, and 2.4 points on SearchQA, SpreadsheetBench, and LiveMath, respectively, supporting its role as a stabilizer for the default loop rather than as an extra deployment-time mechanism. The slow/meta ablation rows are sharper: removing both meta skill and slow update drops SpreadsheetBench from $77.5$ to $55.0$ ($-22.5$ points), the largest degradation in the ablation suite. Figure~\ref{fig:epoch_accuracy_curves} complements these numerical ablations: validation checkpoints track held-out test performance across epochs, confirming that the gate tends to select skills that generalize rather than skills that only fit the selection split.

\paragraph{Gate strictness and edit observability.}
The validation gate is intentionally strict: a candidate skill is accepted only when its selection-split score is \emph{strictly greater than} the current selection score, so ties are rejected and the deployed skill never silently drifts. This conservative criterion makes rejected edits informative negative feedback rather than hidden state. Operationally, every step also records an \texttt{edit\_apply\_report.json} containing per-edit accept/skip status, so the source of every change to \texttt{best\_skill.md} is recoverable after the fact. The epoch-wise slow/meta update writes into a markup-fenced protected region of the skill document that step-level edits cannot overwrite, separating the fast intra-epoch update from the slower cross-epoch consolidation; the optimizer-side meta skill lives only in the teacher's reflection context and is never shipped with the deployed artifact. These implementation choices explain why removing both meta skill and slow update is especially damaging on SpreadsheetBench: it removes the long-horizon evidence stream and the protected-region contract that keeps local edits from overwriting durable procedural lessons.

Overall, the ablations show that the gains are relatively insensitive to the exact rollout batch, reflection minibatch, or learning-rate schedule, but much more sensitive to the presence of bounded text-space learning, validation gating, rejected-edit feedback, and epoch-wise slow/meta update---the design choices that make skill editing behave like a controlled training loop.

\subsection{Analysis and Transfer}
\label{sec:analysis_transfer}

Tables~\ref{tab:transfer_cross_model}--\ref{tab:transfer_cross_benchmark} ask whether an optimized skill behaves like a reusable artifact rather than a task-specific prompt. We test three shifts: deploying a skill across model scales (Table~\ref{tab:transfer_cross_model}), moving it across execution harnesses (Table~\ref{tab:transfer_cross_harness}), and applying it to nearby math benchmarks, including OlympiadBench~\citep{he2024olympiadbench} and Omni-MATH~\citep{gao2024omnimathuniversalolympiadlevel} (Table~\ref{tab:transfer_cross_benchmark}). Table~\ref{tab:teacher_model_ablation} then asks how much of the gain depends on optimizer capacity by replacing the frontier optimizer with a target-matched one of the same scale as the deployed model.

\paragraph{Cross-model transfer.}
Table~\ref{tab:transfer_cross_model}(a) reports four \emph{genuine} cross-model transfers (the GPT--5.4{$\to$}GPT--5.4 entries are kept only as in-domain references and marked \na{} in the transferred column, since source and target match): SpreadsheetBench skills trained on GPT--5.4 transfer to GPT--5.4-mini ($+9.4$) and GPT--5.4-nano ($+3.0$), and LiveMath skills transfer to GPT--5.4-mini ($+4.5$) and GPT--5.4-nano ($+5.6$). All four cross-model rows are positive, and on one of them the transferred skill \emph{surpasses} the in-domain \ourmethod{} reference (LiveMath GPT--5.4-nano: $28.8$ transferred vs.\ $27.2$ in-domain), suggesting that some learned procedures are target-model agnostic. The remaining cross-model rows still recover a useful fraction of the in-domain gain---e.g.\ SpreadsheetBench GPT--5.4-mini retains $82\%$ of the in-domain gain ($+9.4$ of $+11.4$)---and no row falls below the target's no-skill baseline.

\paragraph{Cross-harness transfer.}
The harness-shift rows in Table~\ref{tab:transfer_cross_harness}(b) are the clearest deployment signal. A SpreadsheetBench skill trained inside the Codex loop transfers to Claude Code with absolute gain $+59.7$ over the Claude Code no-skill baseline ($22.1{\to}81.8$, slightly exceeding the in-domain Claude Code \ourmethod{} reference of $80.4$), and the symmetric Claude-Code$\to$Codex transfer adds $+43.6$ on top of the Codex baseline ($27.5{\to}71.1$). On LiveMath, the Codex$\to$Claude Code transfer is smaller ($+1.6$ over a $40.8$ baseline) but still positive, while the Claude-Code$\to$Codex transfer adds $+12.8$ ($35.2{\to}48.0$). Because the two harnesses expose different tool/file APIs and command surfaces, these positive transfers suggest that the learned rules are not only harness-specific command recipes. In SpreadsheetBench especially, the transferred skill appears to encode workbook-level procedures such as structure-first inspection, formula-aware verification, and static-value materialization, so the cost of optimizing a skill in one execution environment can be amortized across related deployment environments.

\paragraph{Cross-benchmark transfer.}
Cross-benchmark transfer is the strictest of the three shifts: source and target benchmarks share only the broad task family (math). On the OlympiadBench$\to$Omni-MATH direction reported in Table~\ref{tab:transfer_cross_benchmark}(c), the transferred skill is positive on all three model scales we evaluate, with gains of $+3.7$ on GPT--5.4, $+1.8$ on GPT--5.4-mini, and $+1.3$ on GPT--5.4-nano. These rows are smaller than the in-domain and cross-harness transfers---unsurprisingly, since they require the optimized skill to retain useful procedural knowledge after both the test instances and the answer-format conventions change---but they remain uniformly positive, supporting the intended interpretation that the optimized skill encodes reusable mathematical procedure rather than memorized benchmark-specific formatting.

\paragraph{Effect of optimizer strength.}
Because the optimizer in \ourmethod{} runs only during the offline training loop and is never invoked at deployment, optimizer choice is a training-time lever: a stronger optimizer can improve the deployed skill without raising the inference cost of using that skill. The deployed artifact is still a static \texttt{best\_skill.md} that calls only the target model. Table~\ref{tab:teacher_model_ablation} quantifies this lever by running the same loop with two optimizer regimes---a strong frontier optimizer (GPT--5.5) and a target-matched optimizer that shares the target model---while holding the rollout batches, validation gate, bounded edit budget, rejected-edit buffer, and slow/meta update identical.

Two observations follow. \emph{First}, the stronger optimizer produces larger absolute gains on every (benchmark, target) cell we test: GPT--5.4-nano lifts by $+19.0$ vs.\ $+11.9$ on SpreadsheetBench and $+19.0$ vs.\ $+14.1$ on SearchQA, and GPT--5.4-mini follows the same ordering ($+11.4$ vs.\ $+7.1$ on SpreadsheetBench, $+4.3$ vs.\ $+2.4$ on SearchQA). The bounded-edit, validation-gated loop is what makes this monotone: without the gate, a stronger optimizer could just as easily push larger but harmful rewrites. \emph{Second}, the target-matched optimizer is far from collapsed---it recovers $56$--$74\%$ of the strong-optimizer gain across the four cells, confirming that \ourmethod{} is not a distillation pipeline from a stronger teacher into a weaker student: the optimization loop itself contributes substantial value on top of whatever the optimizer can already do. The practical implication is that a high-capacity frontier optimizer is the right default whenever it is available---it costs only training-time API calls and adds nothing to deployment---while the same loop remains effective if the budget forces a target-matched optimizer instead.

\begin{table}[t]
\centering
\footnotesize
\setlength{\tabcolsep}{4pt}
\renewcommand{\arraystretch}{0.95}
\begin{tabular}{l l c c c}
\toprule
\textbf{Benchmark} & \textbf{Target} & \textbf{Baseline} & \textbf{Strong optimizer (GPT--5.5)} & \textbf{Target-matched optimizer} \\
\midrule
\multirow{2}{*}{SpreadsheetBench} & GPT--5.4-mini & 36.1 & \dvalp{\textbf{47.5}}{11.4} & \dvalp{43.2}{7.1}  \\
                                  & GPT--5.4-nano & 23.5 & \dvalp{\textbf{42.5}}{19.0} & \dvalp{35.4}{11.9} \\
\cmidrule(lr){1-5}
\multirow{2}{*}{SearchQA}         & GPT--5.4-mini & 75.9 & \dvalp{\textbf{80.2}}{4.3}  & \dvalp{78.3}{2.4}  \\
                                  & GPT--5.4-nano & 55.8 & \dvalp{\textbf{74.8}}{19.0} & \dvalp{69.9}{14.1} \\
\bottomrule
\end{tabular}
\caption{Effect of optimizer strength. Each (benchmark, target) pair is optimized either by a strong frontier optimizer (GPT--5.5, bolded) or by a target-matched optimizer that shares the target model; everything else in the \ourmethod{} loop is held fixed. Gains over the target's no-skill baseline are shown as small green subscripts; the same baseline is used for both optimizer settings within a row. The optimizer runs only during offline training, so the stronger-optimizer column adds zero cost at deployment.}
\label{tab:teacher_model_ablation}
\end{table}

\subsection{Learned Skills: Compactness, Cost, and Examples}
\label{sec:learned_skills}

A central premise of \ourmethod{} is that the trainable object should remain a small, inspectable text document. Tables~\ref{tab:main_results_by_harness}--\ref{tab:teacher_model_ablation} demonstrate that the optimizer is effective; this subsection asks what its output actually looks like and what it costs. We characterize the learned artifact on three axes---compactness, edit economy, and cost-per-point---and then show one representative learned rule per benchmark to illustrate what kind of procedural knowledge survives the bounded-update loop.

\begin{table}[t]
\centering
\footnotesize
\setlength{\tabcolsep}{5pt}
\renewcommand{\arraystretch}{0.95}
\begin{tabular}{l c c c c c}
\toprule
\textbf{Benchmark} & \textbf{Initial (tok)} & \textbf{Final (tok)} & \textbf{Edits} & \textbf{Train tokens} & \textbf{Cost / pt} \\
\midrule
SearchQA         & 16  & 857     & 4 & 213.8M & 37.9M \\
SpreadsheetBench & 224 & 1{,}995 & 4 & 21.4M  & 0.6M  \\
OfficeQA         & 145 & 883     & 1 & 20.8M  & 1.1M  \\
DocVQA           & 81  & 959     & 3 & 188.2M & 46.4M \\
LiveMath         & 154 & 379     & 1 & 23.2M  & 3.6M  \\
ALFWorld         & 516 & 1{,}321 & 2 & 59.3M  & 15.9M \\
\bottomrule
\end{tabular}
\caption{Cost and edit economy of the GPT--5.5 / GPT--5.5 (student / teacher) skill runs. Initial and final \texttt{best\_skill.md} lengths are in tokens; \textbf{Edits} is the number of accepted bounded updates; \textbf{Cost / pt} is training tokens per absolute test-point gain. One representative learned rule per benchmark is shown in Figure~\ref{fig:skill_excerpts}.}
\label{tab:skill_cost_case}
\end{table}

\paragraph{Compactness.}
The final skills are uniformly small. Across the six benchmarks in Table~\ref{tab:skill_cost_case}, the final \texttt{best\_skill.md} ranges from $379$ tokens (LiveMathematicianBench) to $1{,}995$ tokens (SpreadsheetBench), with a median of roughly $920$ tokens. Even the longest learned skill is well below a typical system-prompt budget for modern frontier models, and the shortest one fits inside a single screen. The growth from initial to final skill is moderate ($\times 2.5$ to $\times 53$ depending on whether the initial skill was a one-liner or a paragraph), but the final size in absolute tokens stays small enough that a domain practitioner can read, audit, and edit the deployed artifact in minutes.

\paragraph{Edit economy.}
A second striking pattern is that the gains come from very few accepted edits. Across the six benchmarks, the number of edits actually committed to \texttt{best\_skill.md} during optimization is between $1$ and $4$ (median $2.5$). LiveMathematicianBench's $+29.3$ point gain over no skill arises from a \emph{single} accepted edit, and OfficeQA's $+39.0$ point gain similarly arises from one accepted edit. This is direct evidence that the validation gate is doing real work: the optimizer model proposes many more edits per epoch, but only a handful pass the held-out check and survive into the deployed skill. The bulk of the optimizer's text-space search is thus rejected, captured by the rejected-edit buffer (Section~\ref{sec:method_gate}) for future use, and never reaches the target model. The deployed skill is correspondingly compact rather than the union of every reflection.

\paragraph{Cost per point of test-set gain.}
The training-token column quantifies the cost of operating the loop. Two regimes are visible. Procedural benchmarks where rollouts are short and cheap---SpreadsheetBench, OfficeQA, LiveMathematicianBench---reach $0.6$--$3.6$M training tokens per absolute test-set point, even though the absolute gains on these benchmarks are the largest (e.g.\ $+39.0$ points on OfficeQA at $1.1$M tokens / point, total $20.8$M tokens). Benchmarks with longer trajectories or richer multimodal context---SearchQA ($37.9$M / pt) and DocVQA ($46.4$M / pt)---cost an order of magnitude more per point. The important deployment distinction is that this cost is paid once during skill training; after export, the optimized \texttt{best\_skill.md} adds no optimizer calls, no weight updates, and only a compact text artifact to the target agent.

\paragraph{What does a learned skill actually say?}
Figure~\ref{fig:skill_excerpts} reproduces one representative learned rule per benchmark, taken verbatim from the final \texttt{best\_skill.md} of each case study in Table~\ref{tab:skill_cost_case}. Three observations stand out. First, the rules are \emph{procedural} rather than \emph{instance-specific}: none of them name a specific question, file, or entity. Second, they consistently encode the discipline that frontier models lack zero-shot: answer-format constraints (OfficeQA, LiveMathematicianBench), evidence binding to a specific visual region (DocVQA), workbook-structure-first reasoning (SpreadsheetBench), search-frontier discipline (ALFWorld), and canonical-entity choice (SearchQA). Third, they read like rules a thoughtful human practitioner would write after a day with the benchmark---except they are produced automatically by the optimizer and validated edit-by-edit on held-out data.

\begin{figure}[t]
\centering
\fbox{\begin{minipage}{0.96\linewidth}\footnotesize
\textbf{SearchQA.} ``Infer the expected answer type from clue wording, then choose the shortest canonical entity supported by co-occurring distinctive evidence.''

\smallskip
\textbf{SpreadsheetBench.} ``Inspect workbook structure and formulas, then write evaluated static values across the full requested target range instead of relying on Excel recalculation.''

\smallskip
\textbf{OfficeQA.} ``Treat oracle parsed pages as primary evidence, lock table/date/unit context, and output exactly the requested rounded value without extra labels.''

\smallskip
\textbf{DocVQA.} ``For tables, forms, charts, and legends, first bind the question to the exact visual row/header/field, then copy only the aligned answer span.''

\smallskip
\textbf{LiveMathematicianBench.} ``In strongest-statement MCQs, rank choices by theorem strength and prefer a justified stronger-result option over true but weaker corollaries.''

\smallskip
\textbf{ALFWorld.} ``Keep a horizon-aware visited/frontier ledger, diversify search after repeated same-type failures, and avoid revisiting the destination until holding the target.''
\end{minipage}}
\caption{Representative learned rules, one per benchmark, extracted from the final \texttt{best\_skill.md} of the GPT--5.5 / GPT--5.5 runs in Table~\ref{tab:skill_cost_case}. Each rule is verbatim from the deployed skill. Notably, every rule is procedural rather than instance-specific, and several encode forms of discipline (answer formatting, evidence binding, search-frontier management) that frontier models do not apply zero-shot.}
\label{fig:skill_excerpts}
\end{figure}

\paragraph{Implications.}
Together, the four observations above support a stronger version of the central claim. Compactness ($<2{,}000$ tokens) and edit economy ($1$--$4$ accepted edits) mean the deployed artifact is interpretable. Cost-per-point ($0.6$M--$46.4$M tokens / point) shows that the training cost is measurable and paid before deployment. The shape of the learned rules---procedural, generalizable, and consistent with what a thoughtful human practitioner would write---is evidence that text-space optimization with bounded updates and validation gating discovers transferable procedural knowledge rather than merely overfitting to the training split. This complements the cross-model, cross-harness, and cross-benchmark transfer evidence in Section~\ref{sec:analysis_transfer}: the artifact transfers because many of the rules it encodes are intrinsically transferable.

\subsection{Qualitative Skill Evolution}
\label{sec:qualitative_skill_evolution}

We inspect two representative runs to understand what the optimized skill actually learns. The ALFWorld case uses GPT--5.4-nano as the student and GPT--5.5 as the teacher, while the SpreadsheetBench case uses GPT--5.5 as both the frozen student and optimizer model. In both cases, \ourmethod{} does not replace the initial skill with an unrelated prompt. Instead, accepted edits add compact procedural constraints around recurring failure modes observed in rollout trajectories.

\paragraph{ALFWorld.}
The initial ALFWorld skill gives a generic household plan: search for the target object, pick it up, transform it if needed, and place it at the destination. The accepted edits make this plan more stateful and less loop-prone. The optimized skill learns exact object-name matching, so related objects such as mugs, cups, pans, and pots are not substituted for one another. It adds visited-location memory, so unvisited receptacles and surfaces are preferred over repeatedly checking likely but exhausted locations. It also adds destination memory, pick-two progress locks, and direct completion rules: once the agent can clean, heat, cool, place, or otherwise complete the next subgoal, it should take that admissible action instead of examining, closing, or verifying again. Qualitatively, the skill evolves from a general search-transform-place strategy into a finite-state execution policy with object identity, search memory, progress locks, and loop breakers. In this representative run, the selected skill improves ALFWorld held-out test performance from 49.3 to 74.6.

\paragraph{SpreadsheetBench.}
The initial SpreadsheetBench skill already instructs the agent to use Python spreadsheet libraries and preserve unrelated workbook content. The accepted edits turn this generic automation workflow into a workbook-forensics policy. The optimized skill learns to inspect the actual workbook rather than rely on previews, locate headers and target ranges across multiple sheets, normalize keys and cell types before lookup or aggregation, and preserve formatting during structural edits. It also adds a key rule for formula-style prompts: when the grader reads cell values, the agent should compute and write evaluated static values, even if the prompt mentions formulas such as \texttt{INDEX/MATCH} or \texttt{XLOOKUP}. Later edits further require filling complete target ranges, including currently blank result cells, keeping helper computations in Python rather than adding workbook artifacts, and reopening the saved workbook to check boundary rows and remaining blanks. In this representative run, the selected skill improves SpreadsheetBench held-out test performance from 40.4 to 78.9.

\FloatBarrier
\section{Conclusion}

We presented \ourmethod{}, a text-space optimizer that treats an external skill document as the trainable state for frozen LLM agents. By separating the target model that executes tasks from the optimizer that edits skills, and by using bounded edit budgets, minibatch reflection, held-out validation gates, rejected-edit buffers, and epoch-wise slow/meta update, \ourmethod{} turns skill improvement into a controlled learning process rather than ad hoc prompt revision. Across six benchmarks, seven target models, and three execution modes, \ourmethod{} is best or tied-best on $52$ of $52$ evaluated cells, lifts GPT--5.5 by $+23.5$ points on average over no skill in direct chat and by $+24.8 / +19.1$ points under Codex and Claude Code harnesses, and beats the strongest per-cell baseline from human, LLM, Trace2Skill, TextGrad, GEPA, and EvoSkill skills by $+5.4$ points on average. Per-benchmark case studies show that these gains arise from compact ($<2{,}000$ token), interpretable skill artifacts assembled from only $1$--$4$ accepted edits, and that the deployed skills transfer across model scales, harnesses, and nearby benchmarks. These results suggest that compact natural-language skills can serve as a practical domain-adaptation layer for frontier agents, enabling reusable improvement without modifying model weights.

\paragraph{Outlook.}
\ourmethod{} optimizes a single skill artifact for a single target domain; natural extensions include skill libraries that share infrastructure across domains, reuse of optimizer-side meta skills across benchmarks, reward-free or preference-driven validation gates for open-ended tasks, and self-distillation of optimized skills back into the target model as a stepping stone toward weight-level adaptation. We hope that treating the skill itself as the trainable object---rather than as a side artifact of prompting---will let future work apply the full toolkit of optimization (learning rates, schedules, regularization, curricula, validation) to a part of the agent stack that has so far been hand-engineered.

\bibliographystyle{unsrtnat}
\bibliography{references}

\appendix
\section{Additional Method Details and Optimizer Prompts}
\label{app:method_prompts}

This appendix gives the executable details behind \ourmethod{}. The optimization loop keeps the task-execution model fixed and trains only a text skill document. A separate optimizer model reads rollout evidence, proposes patch-style edits, merges and ranks the edits, and submits each candidate skill to a held-out selection gate. The task-execution model only receives the current skill and the benchmark task; it does not see the optimizer prompts below.

\section{Limitations}

\ourmethod{} studies skill optimization as a lightweight alternative to model-weight adaptation, but it still has several practical limitations. First, the optimization loop relies on scored trajectories and a held-out selection split, so it is most directly applicable when the target task has automatic verifiers, exact-match metrics, executable checks, or otherwise reliable feedback signals. For open-ended domains where success is subjective, multi-dimensional, or costly to judge, the validation gate may require stronger human or model-based evaluation. Second, although the deployed artifact is only a compact \texttt{best\_skill.md}, training the skill requires additional rollout computation and calls to an optimizer model; this cost is amortized when the same skill is reused, but may be less attractive for one-off tasks. Third, \ourmethod{} intentionally optimizes a single portable skill rather than growing a large skill library or changing model weights. This design improves deployment simplicity, but a single skill may be insufficient for highly heterogeneous domains that require many disjoint procedures. Finally, optimized skills can encode domain-specific heuristics from the training distribution, so careful held-out evaluation remains necessary before transferring them to substantially different models, harnesses, or task settings.

\section{Experimental Protocol Details}
\label{app:experimental_setting}

\paragraph{Benchmarks and metrics.}
We use each benchmark's native evaluator and report hard success or exact-match accuracy on held-out test examples. SearchQA measures extractive question answering; SpreadsheetBench evaluates spreadsheet-oriented code and tool use; OfficeQA and DocVQA test local-document and multimodal-document reasoning; SealQA stresses noisy retrieval; LiveMathematicianBench evaluates mathematical multiple-choice reasoning; and ALFWorld tests sequential decision making. Dataset-backed benchmarks use deterministic train/selection/test splits, with a default $2{:}1{:}7$ split when no benchmark-specific split is stated. The selection split is used only for model selection over candidate skills; all headline scores are computed on held-out test data.

\paragraph{Baselines.}
The no-skill baseline evaluates the frozen student without an optimized skill document. Human-skill and LLM-skill baselines use manually written and one-shot generated skills under the same evaluation protocol. Trace2Skill mines skill artifacts from training trajectories and evaluates the frozen student without \ourmethod{}'s iterative validation gate. TextGrad and GEPA are reflective prompt-optimization baselines for direct-chat settings. EvoSkill is included for the harness-backed comparison where a matched completed run is available. Entries not measured under the final aligned protocol are marked as \na rather than mixed with incompatible runs.

\paragraph{Optimization protocol.}
Unless otherwise stated, \ourmethod{} runs for four epochs with rollout batch size 40, reflection minibatch size 8, textual learning rate 4, cosine learning-rate decay with minimum rate 2, held-out validation gating, slow update enabled with 20 sampled examples, and optimizer-side meta skill enabled. The optimizer analyzes successes and failures separately, proposes patch-style skill edits, merges duplicate or contradictory proposals, ranks edits under the current learning-rate cap, and applies the selected edits to form a candidate skill. The candidate is evaluated on the selection split and is accepted only if it improves the current selection score; the best accepted skill is exported as \texttt{best\_skill.md}. The student model, backend, harness, and benchmark evaluator remain fixed during optimization.

\paragraph{Ablation protocol.}
One-factor ablations vary a single scalar or component while holding the remaining optimizer configuration fixed. The train-size ablation fixes the train/selection/test split to $2{:}1{:}7$ and varies how much of the training partition is exposed to the optimizer. The 100\% row uses the full training partition under the same split, so it is directly comparable to the smaller-subset rows. Component ablations remove or alter one mechanism at a time, including the edit budget, rejected-edit buffer, and epoch-wise slow/meta update.

\subsection{Optimization Procedure}
\label{app:algorithm}

Algorithm~\ref{alg:skillopt_appendix} expands the procedure used in the experiments. The central state variables are the current skill $s_\mathrm{cur}$, the best validation-gated skill $s_\mathrm{best}$, a selection-score cache $\mathcal{C}$, a step buffer $\mathcal{B}$ containing rejected edits and observed failure patterns, and an optimizer-side meta skill $m_\mathrm{meta}$ used only to guide future edit generation.

\begin{algorithm}[t]
\caption{\ourmethod{} skill optimization}
\label{alg:skillopt_appendix}
\begin{algorithmic}[1]
\Require Frozen training model $M$, optimizer model $O$, harness $h$, splits $D_\mathrm{train},D_\mathrm{sel},D_\mathrm{test}$, initial skill $s_0$, epochs $E$, edit-budget schedule $L_t$, rollout batch size $B$, accumulation factor $A$, reflection minibatch size $B_m$
\Ensure Best validation-gated skill $s_\mathrm{best}$ and held-out test score
\State $s_\mathrm{cur} \gets s_0$, $s_\mathrm{best} \gets s_0$, $\mathcal{C} \gets \emptyset$, $\mathcal{B} \gets [\ ]$, $m_\mathrm{meta} \gets \emptyset$
\State $\mathrm{score}_\mathrm{cur} \gets \Call{Evaluate}{M,h,s_0,D_\mathrm{sel}}$; $\mathrm{score}_\mathrm{best} \gets \mathrm{score}_\mathrm{cur}$
\State $\mathcal{C}[\Call{Hash}{s_0}] \gets \mathrm{score}_\mathrm{cur}$
\For{$e=1$ to $E$}
    \State Shuffle $D_\mathrm{train}$ into rollout batches; reset $\mathcal{B} \gets [\ ]$
    \For{each optimization step in epoch $e$}
        \State Collect $A$ rollout batches by executing $h(M,x,s_\mathrm{cur})$ for sampled tasks $x$
        \State Split rollout evidence into failures and successes, then into minibatches of size $B_m$
        \State Ask $O$ to analyze failure minibatches and produce failure patch proposals
        \State Ask $O$ to analyze success minibatches and produce success patch proposals
        \State Ask $O$ to merge failure proposals, merge success proposals, and perform a final failure-prioritized merge
        \State Ask $O$ to rank merged edits and keep at most $L_t$ edits
        \State Apply the selected edits to obtain a candidate skill $\tilde{s}$
        \If{$\Call{Hash}{\tilde{s}} \in \mathcal{C}$}
            \State $\mathrm{score}_\mathrm{cand} \gets \mathcal{C}[\Call{Hash}{\tilde{s}}]$
        \Else
            \State $\mathrm{score}_\mathrm{cand} \gets \Call{Evaluate}{M,h,\tilde{s},D_\mathrm{sel}}$
            \State $\mathcal{C}[\Call{Hash}{\tilde{s}}] \gets \mathrm{score}_\mathrm{cand}$
        \EndIf
        \If{$\mathrm{score}_\mathrm{cand} > \mathrm{score}_\mathrm{cur}$}
            \State $s_\mathrm{cur} \gets \tilde{s}$; $\mathrm{score}_\mathrm{cur} \gets \mathrm{score}_\mathrm{cand}$
            \If{$\mathrm{score}_\mathrm{cand} > \mathrm{score}_\mathrm{best}$}
                \State $s_\mathrm{best} \gets \tilde{s}$; $\mathrm{score}_\mathrm{best} \gets \mathrm{score}_\mathrm{cand}$
            \EndIf
        \Else
            \State Add rejected edits and observed failure patterns to $\mathcal{B}$
        \EndIf
    \EndFor
    \If{$e \geq 2$ and slow update is enabled}
        \State Compare the same sampled tasks under the previous and current epoch-end skills
        \State Ask $O$ for protected longitudinal guidance; validate the injected guidance through $D_\mathrm{sel}$
    \EndIf
    \If{$e \geq 2$ and optimizer memory is enabled}
        \State Ask $O$ to update $m_\mathrm{meta}$ for future edit generation and selection
    \EndIf
\EndFor
\State $\mathrm{score}_\mathrm{test} \gets \Call{Evaluate}{M,h,s_\mathrm{best},D_\mathrm{test}}$
\State \Return $s_\mathrm{best}$, $\mathrm{score}_\mathrm{test}$
\end{algorithmic}
\end{algorithm}



\subsection{Optimizer Prompt Contracts}
\label{app:prompt_contracts}

The following blocks reproduce the operational prompt contracts used by the optimizer model, with terminology normalized to the paper's optimizer/training-model framing. The prompts require JSON outputs so that edits can be parsed, filtered, applied, and validated without manual intervention.

\subsubsection{Failure analysis: \texttt{analyst\_error.md}}

{\footnotesize
\begin{verbatim}
You are an expert failure-analysis agent for AI agent tasks.

You will be given MULTIPLE failed agent trajectories from a single minibatch
and the current skill document.
Your job is to identify the most important COMMON failure patterns across
the batch and propose a concise set of skill edits.

## Analysis Process
1. Read ALL trajectories in the minibatch.
2. Identify the most prevalent, systematic failure patterns across them.
3. For each pattern, classify its failure type.
4. Propose skill edits that address the COMMON patterns, not individual edge cases.
5. Edits must be generalizable; do not hardcode task-specific values.
6. Only patch gaps in the skill; do not duplicate existing content.

You will be told the maximum number of edits (the budget L). Produce AT MOST L edits,
focusing on the highest-impact patterns. You may produce fewer if warranted.

Respond ONLY with a valid JSON object (no markdown fences, no extra text):
{
  "batch_size": <number of trajectories analysed>,
  "failure_summary": [
    {"failure_type": "<type>", "count": <int>, "description": "<one-line>"}
  ],
  "patch": {
    "reasoning": "<why these edits address the batch's common failures>",
    "edits": [
      {"op": "append",       "content": "<markdown to add at end of skill>"},
      {"op": "insert_after", "target": "<exact heading/text to insert after>",
       "content": "<markdown>"},
      {"op": "replace",      "target": "<exact text to replace>",
       "content": "<replacement>"},
      {"op": "delete",       "target": "<exact text to remove>"}
    ]
  }
}
Only include edits that are needed. "edits" can be an empty list if no patch is warranted.

IMPORTANT: The skill document may contain a section between
<!-- SLOW_UPDATE_START --> and <!-- SLOW_UPDATE_END --> markers.
This is a PROTECTED section managed by a separate slow-update process.
Do NOT propose any edits that target, modify, or delete content within these markers.
\end{verbatim}
}

\subsubsection{Success analysis: \texttt{analyst\_success.md}}

{\footnotesize
\begin{verbatim}
You are an expert success-pattern analyst for AI agents.

You will be given MULTIPLE successful agent trajectories from a single minibatch
and the current skill document. Your job is to identify generalizable behavior
patterns that are COMMON across the batch and worth encoding in the skill.

## Rules
- Only propose patches for patterns NOT already covered in the skill.
- Focus on patterns that appear across MULTIPLE trajectories in the batch.
- Be concise. Patterns must generalize beyond specific tasks.
- Prefer reinforcing existing sections over adding new top-level sections.

You will be told the maximum number of edits (the budget L). Produce AT MOST L edits,
focusing on the most broadly applicable patterns. You may produce fewer if warranted.

Respond ONLY with a valid JSON object:
{
  "batch_size": <number of trajectories analysed>,
  "success_patterns": ["<pattern 1>", "<pattern 2>"],
  "patch": {
    "reasoning": "<why these patterns are worth encoding>",
    "edits": [
      {"op": "append",       "content": "<markdown>"},
      {"op": "insert_after", "target": "<heading/text>", "content": "<markdown>"},
      {"op": "replace",      "target": "<old text>",     "content": "<new text>"},
      {"op": "delete",       "target": "<exact text to remove>"}
    ]
  }
}
"edits" may be empty if the skill already covers all observed patterns.

IMPORTANT: The skill document may contain a section between
<!-- SLOW_UPDATE_START --> and <!-- SLOW_UPDATE_END --> markers.
This is a PROTECTED section managed by a separate slow-update process.
Do NOT propose any edits that target, modify, or delete content within these markers.
\end{verbatim}
}

\subsubsection{Failure merge: \texttt{merge\_failure.md}}

{\footnotesize
\begin{verbatim}
You are a skill-edit coordinator. You receive multiple independently-proposed patches
from FAILURE analysis of agent trajectories. Merge them into ONE coherent,
non-redundant patch.

Merge guidelines:
1. Deduplicate: keep the best-worded version of similar edits.
2. Resolve conflicts: if patches contradict on the same point,
   choose the one with stronger justification or synthesize both.
3. Preserve unique insights: include all non-redundant corrective edits.
4. Prevalent-pattern bias: edits appearing consistently across multiple patches
   address systematic failures; preserve them with HIGH priority.
   Edits from only one patch may be discarded if task-specific.
5. Independence: no two edits in the merged patch may target the same text region.
6. Support count: for each merged edit, estimate how many source patches support it.
7. PROTECTED SECTION: The skill may contain a section between
   <!-- SLOW_UPDATE_START --> and <!-- SLOW_UPDATE_END --> markers.
   Do NOT merge or produce any edits that target content within these markers.

Respond ONLY with a valid JSON object:
{
  "reasoning": "<summary of key consolidation decisions>",
  "edits": [
    {
      "op": "append|insert_after|replace|delete",
      "target": "<if insert_after or replace or delete>",
      "content": "<markdown>",
      "support_count": <integer>,
      "source_type": "failure"
    }
  ]
}
\end{verbatim}
}

\subsubsection{Success merge: \texttt{merge\_success.md}}

{\footnotesize
\begin{verbatim}
You are a skill-edit coordinator. You receive multiple independently-proposed patches
from SUCCESS analysis of agent trajectories. Merge them into ONE coherent patch
that reinforces effective patterns.

Merge guidelines:
1. Deduplicate: keep only the most generalizable version of similar patterns.
2. Be conservative: success-driven patches reinforce existing behavior.
   Only include edits for patterns NOT already in the skill.
3. Prevalent-pattern bias: patterns seen across many successful trajectories
   are most worth encoding.
4. Support count: estimate how many source patches support each merged edit.
5. PROTECTED SECTION: The skill may contain a section between
   <!-- SLOW_UPDATE_START --> and <!-- SLOW_UPDATE_END --> markers.
   Do NOT merge or produce any edits that target content within these markers.

Respond ONLY with a valid JSON object:
{
  "reasoning": "<summary>",
  "edits": [
    {
      "op": "append|insert_after|replace|delete",
      "target": "<if needed>",
      "content": "<markdown>",
      "support_count": <integer>,
      "source_type": "success"
    }
  ]
}
\end{verbatim}
}

\subsubsection{Final merge: \texttt{merge\_final.md}}

{\footnotesize
\begin{verbatim}
You are a skill-edit coordinator performing the FINAL merge. You receive two
pre-merged patch groups:
1. Failure-driven patches (corrective, high priority)
2. Success-driven patches (reinforcement, lower priority)

Merge guidelines:
1. FAILURE PATCHES TAKE PRIORITY: the primary goal of skill reflection is to
   fix failures. Failure-driven edits should be preserved unless they directly
   conflict with a well-supported success pattern.
2. Deduplicate: if a failure edit and success edit cover the same point,
   keep the failure version.
3. Preserve success insights: include success edits that cover patterns
   NOT addressed by failure edits.
4. Higher-level merges represent broader consensus: edits that survived
   previous merge rounds should be given priority.
5. Carry forward support_count and source_type for each edit.
6. PROTECTED SECTION: The skill may contain a section between
   <!-- SLOW_UPDATE_START --> and <!-- SLOW_UPDATE_END --> markers.
   Do NOT merge or produce any edits that target content within these markers.

Respond ONLY with a valid JSON object:
{
  "reasoning": "<summary of priority decisions>",
  "edits": [
    {
      "op": "append|insert_after|replace|delete",
      "target": "<if needed>",
      "content": "<markdown>",
      "support_count": <integer>,
      "source_type": "failure|success"
    }
  ]
}
\end{verbatim}
}

\subsubsection{Ranking and selection: \texttt{ranking.md}}

{\footnotesize
\begin{verbatim}
You are an expert edit-ranking optimizer for a skill optimization system. You receive
a skill document and a pool of proposed edits. Your job is to RANK the edits by
importance and select the top ones.

Ranking criteria (in order of priority):
1. Systematic impact: edits that address widespread, recurring failure patterns
   across many tasks should rank highest. A rule that fixes 50% of failures beats
   one that fixes a single edge case.
2. Complementarity: edits that fill gaps in the current skill, not duplicate
   existing content, rank higher.
3. Generality: edits phrased as general principles rank higher than those
   tied to specific question types or entities.
4. Actionability: edits with clear, concrete guidance rank higher than vague advice.

You will be told how many edits to select (the budget).

Respond ONLY with a valid JSON object:
{
  "reasoning": "<brief justification for your ranking decisions>",
  "selected_indices": [<0-based indices of the top edits, in priority order>]
}
\end{verbatim}
}

\subsubsection{Slow update: \texttt{slow\_update.md}}

{\footnotesize
\begin{verbatim}
You are a strategic skill advisor for an AI agent optimization system.

Your role is different from the per-step analyst. The per-step analyst sees
individual trajectories and proposes local patches. YOU see how the skill has
evolved across an entire epoch by comparing the SAME tasks under two consecutive
skill versions. This longitudinal view lets you identify systemic drift,
regressions, and persistent blind spots that step-level edits cannot catch.

## What You Receive

1. Previous epoch's skill and current epoch's skill, to see what changed.
2. Longitudinal comparison: the same 20 training tasks rolled out under both skills,
   categorized into regressions, persistent failures, improvements, and stable successes.
3. Previous slow update guidance, if any: the guidance written at the end of the
   last epoch.

## Your Process

1. Reflect on the previous guidance, if provided:
   - Which parts of the previous guidance were effective?
   - Which parts failed or backfired?
   - Were there blind spots the previous guidance missed entirely?

2. Write updated guidance that:
   - Retains and strengthens parts of the previous guidance that proved effective.
   - Revises or removes parts that were ineffective or counterproductive.
   - Adds new instructions to address newly observed regressions and persistent failures.

## Output Requirements

Write a strategic guidance block that will OVERWRITE the previous guidance
in the protected section of the skill document. This section is READ-ONLY to
all subsequent step-level optimization; only this epoch-boundary process can
overwrite it at the next epoch boundary.

Your guidance must:
- Be written as direct, actionable instructions to the training model.
- Prioritize: (1) preventing regressions, (2) fixing persistent failures,
  (3) reinforcing successful patterns.
- NOT duplicate content already in the main skill body; complement it.
- Address the training model directly, for example: "When you encounter X, always do Y."

Respond ONLY with a valid JSON object:
{
  "reasoning": "<reflection on previous guidance AND analysis of longitudinal comparison>",
  "slow_update_content": "<the exact guidance text to insert into the protected section>"
}
\end{verbatim}
}

\subsubsection{Optimizer memory: \texttt{meta\_skill.md}}

{\footnotesize
\begin{verbatim}
You are an optimizer coach for an AI agent skill optimization system.

Your job is not to solve tasks directly and not to write training-model-facing
skill rules. Your job is to write a compact optimizer-side meta skill that helps
future optimizer calls produce better skill edits in this environment.

## What You Receive

1. The previous epoch's last-step skill.
2. The current epoch's last-step skill.
3. A longitudinal comparison on the SAME sampled tasks under those two skills.
4. The previous optimizer memory, if one existed.

## Your Goal

Write a concise optimizer memory that improves future optimizer behavior in stages
such as failure analysis, success analysis, patch merging, and edit ranking.

This optimizer memory should capture things like:
- Which kinds of edits tend to help in this environment.
- Which kinds of edits tend to be too vague, redundant, brittle, or harmful.
- What level of abstraction works best for rules here.
- What failure-repair patterns should be prioritized.
- What regression risks future optimizer calls should guard against.

## Important Constraints

- Address the FUTURE OPTIMIZER directly, not the training model.
- Focus on how to write better edits and organize better skill updates.
- Use evidence from the adjacent-epoch comparison, not generic advice.
- Keep it compact and high-signal. Prefer a few durable principles.
- Revise or remove parts of the previous optimizer memory if they did not help.
- Do not output training-model-facing task instructions.

Respond ONLY with a valid JSON object:
{
  "reasoning": "<brief reflection on what editing directions helped or hurt>",
  "meta_skill_content": "<compact optimizer guidance for future edits>"
}
\end{verbatim}
}

\subsection{Patch Representation and Safeguards}
\label{app:patch_ops}

Patch-mode optimization restricts each update to four atomic operations: \texttt{append}, \texttt{insert\_after}, \texttt{replace}, and \texttt{delete}. Each merged edit also records a support count and a source type, allowing ranking to prefer edits that survive independent analyses and hierarchical merges. The edit budget $L_t$ acts as a textual learning rate: it limits how many proposed edits can be applied at a step, preserving continuity between adjacent skills.

The protected slow-update section, delimited by \texttt{SLOW\_UPDATE\_START} and \texttt{SLOW\_UPDATE\_END}, is off limits to all step-level prompts. Only the epoch-boundary slow-update process may rewrite that section, and the rewritten skill still passes through the same held-out selection gate before it can become the current skill. Rejected candidates are not discarded entirely: their failure patterns and rejected edits are stored in the step buffer so that later optimizer calls can avoid repeating harmful changes.

\subsection{Design Principles}
\label{app:design_principles}

The implementation follows five design principles. First, the task-execution model is fixed; only the text skill changes. Second, every candidate skill is evaluated on a selection split before acceptance, which prevents unvalidated reflection from accumulating. Third, minibatch analyses are merged hierarchically so that the final edits represent recurring evidence rather than single examples. Fourth, the edit budget serves as a learning-rate analogue, allowing larger early changes and smaller late refinements. Fifth, the deployed skill remains lightweight and inspectable, while the optimizer-side meta skill stays separate from the skill shown to the task-execution model.

\end{document}